\documentclass[sigconf]{acmart}
\AtBeginDocument{%
  }

\usepackage{booktabs}
\usepackage{multirow}
\usepackage{graphicx}
\usepackage{xcolor} %
\usepackage{array}  %
\usepackage{hyperref}
\usepackage{siunitx} %
\usepackage{dsfont} 
\usepackage{subfigure}
\usepackage{multirow}
\usepackage{array}
\usepackage{makecell}
\usepackage{threeparttable}
\usepackage{graphicx}
\usepackage{textcomp}
\usepackage{xcolor}
\usepackage{enumitem}
\usepackage{comment}
\usepackage{pifont}
\usepackage{amsthm}
\usepackage{algorithm}
\usepackage{amsmath}
\usepackage{bm}
\usepackage{bbding}
\usepackage{amsthm}
\usepackage{algpseudocode}
\usepackage{fontawesome}

\usepackage{amsmath}
\usepackage{amsthm}

\usepackage{dsfont}
\usepackage{colortbl}  %
\usepackage{xcolor}

\renewcommand\footnotetextcopyrightpermission[1]{} %

\begin{document}

\title{GLIDE: Scalable and Event-Aware Inductive Forecasting over Evolving Logistics Graphs}

\author{Zhi Sheng}
\affiliation{%
  \institution{Tsinghua University}
  \city{Beijing}
  \country{China}}
\email{chengz23@mails.tsinghua.edu.cn}
\authornote{Equally contribution.}

\author{Chuyue Wang$^*$}
\affiliation{%
  \institution{Meituan}
  \city{Beijing}
  \country{China}}
\email{chuyuewang0710@gmail.com}

\author{Jiahao Ji$^*$}
\affiliation{%
  \institution{Meituan}
  \city{Beijing}
  \country{China}}
\email{jijiahao03@meituan.com}

\author{Yile Liang}
\affiliation{%
  \institution{Meituan}
  \city{Beijing}
  \country{China}}
\email{yileliang0412@163.com}

\author{Xuetao Ding}
\affiliation{%
  \institution{Meituan}
  \city{Beijing}
  \country{China}}
\email{dingxuetao@meituan.com}

\author{Jinghua Hao}
\affiliation{%
  \institution{Meituan}
  \city{Beijing}
  \country{China}}
\email{haojinghua@meituan.com}

\author{Jie Feng}
\affiliation{%
  \institution{Zhongguancun Academy}
  \city{Beijing}
  \country{China}}
\email{fengj12ee@hotmail.com}

\author{Yong Li}
\affiliation{%
  \institution{Tsinghua University}
  \city{Beijing}
  \country{China}}
\email{liyong07@tsinghua.edu.cn}

\renewcommand{\shortauthors}{Zhi Sheng, Chuyue Wawng, Jiahao Ji et al.}

\begin{abstract}

Accurate forecasting of order volume at scale-effect hotspots (SEHs), spatially coherent groups of logistics flows that serve as distributed optimization units for real-time many-to-one dispatch, is a prerequisite for efficient order pooling in on-demand food delivery (OFD) platforms such as Meituan. However, real-world logistics networks present unique challenges that render standard spatiotemporal prediction model ineffective. (i) Dynamic Topology: Unlike static road networks, logistics graphs are ephemeral, with hotspot emerging and vanishing continuously; (ii) Scale--Volume Heterogeneity: Graph sizes vary from $10^2$ to $10^4$, while zero-inflated, long-tailed volumes create computational bottlenecks and bias predictions toward inactivity; and (iii) Event-Driven Disruption: Exogenous events (holidays, weather, academic calendars) induce systematic demand shifts that purely data-driven backbones cannot anticipate. In this paper, we present GLIDE, a production-ready framework that models evolving logistics graphs in a scalable latent space while adapting to heterogeneous demand regimes and event-driven shifts. GLIDE introduces a Graph-Guided Latent Aggregation mechanism, which dynamically projects varying physical subgraphs into a fixed-size, semantically stable latent space via a differentiable sparse assignment. Furthermore, we design a Volume-Adaptive MoE Head that explicitly disentangles high-volume, bursty demand from low-volume, sparse demand, thereby alleviating long-tail regression bias. Finally, we propose an Event-Aware Modulation module, a lightweight modulator steered by an offline LLM-constructed event knowledge bank that performs targeted multiplicative corrections on a frozen backbone, mitigating event-period bias without retraining. Extensive experiments on real-world datasets and online deployed applications from Meituan demonstrate that GLIDE outperforms SOTA baselines in accuracy, efficiency and usability.
\end{abstract}

\maketitle

\section{Introduction}

The rapid urbanization and digitalization of modern life have propelled OFD services into a critical infrastructure for daily living. Leading platforms like Meituan now process over 70 million orders daily, necessitating dispatch systems capable of matching massive order volumes with couriers in real-time. Such dispatch and route-optimization capabilities echo the broader line of industrial logistics and ETA prediction systems~\cite{derrow2021eta,hu2022deepreta,jin2019coride,garg2025food,liang2023enhancing,liang2024harvesting}. Traditionally, platforms have relied on coarse-grained demand aggregation using static grids or administrative regions. However, the pursuit of operational efficiency now demands granular precision at the logistics flow level. To facilitate optimized courier route generation, recent systems utilize Order Pooling~\cite{lei2025pooling, chen2024order}, which consolidates orders into efficient bundles. As established in recent production deployments, the fundamental atom of these logistics trajectories is the Flow Unit~\cite{liang2024harvesting}, defined as a directed vector connecting a specific Pick-up Area of Interest (AOI) to a Delivery AOI~\cite{liang2024harvesting}. While existing algorithms can optimize assignments for current orders, their efficacy is bounded by the system's capacity to anticipate future logistics density~\cite{liang2023enhancing}. Accurate forecasting of SEH demand is a prerequisite for the skilled couriers delivery network (SCDN)~\cite{liang2024harvesting}, enabling dispatch systems to preemptively identify high-density flows and effectively prune the vast search space for the many-to-one assignment (MoA) problem.

In this paper, we focus on forecasting the order volume of \textbf{scale-effect hotspots (SEHs)}, where each SEH represents the aggregate demand intensity of a specific group of FUs within a discrete time window~\cite{liang2024harvesting}. We model SEHs as nodes in a graph to capture their spatiotemporal correlations, such as shared pick-up AOIs and synchronized peak-hour patterns. However, accurate forecasting in production environments faces three fundamental challenges. \textbf{(i) Evolving Topologies and Inductive Constraints:} Unlike road networks with fixed intersections, logistics networks are open systems in which SEHs continuously emerge and disappear as merchant supply and user preferences evolve. Consequently, the active node set $\mathcal{V}_t$ and its induced subgraph vary substantially over time. Standard transductive models, such as DCRNN~\cite{li2017diffusion} and STGCN~\cite{yu2017spatio}, bind their parameters or representations to fixed node identities and therefore cannot generalize to unseen SEHs without retraining. Although dynamic-graph methods~\cite{pareja2020evolvegcn,you2022roland} partially accommodate topology evolution, they typically assume persistent node identities and remain ineffective when entirely new SEHs emerge in an open-set manner. \textbf{(ii) Scale--Volume Heterogeneity:} Production forecasting operates over graphs whose sizes vary from local micro-zones with approximately $10^2$ active nodes to city-wide networks with approximately $10^4$ nodes. Meanwhile, during the experiments, SEH volumes exhibit a highly imbalanced, zero-inflated, and long-tailed distribution: a head SEH may generate more than 100 orders per hour, whereas a tail SEH may generate fewer than 5. These coupled variations create both computational and statistical challenges. Global-attention models incur quadratic complexity as the number of active nodes increases, making city-wide inference prohibitively expensive, while standard regression objectives are dominated by abundant sparse-demand samples, leading to systematic under-forecasting of high-volume SEHs that are critical to pooling efficiency. \textbf{(iii) Event-Driven Demand Disruption:} Beyond the regularities captured from historical observations, OFD demand is frequently disrupted by exogenous events, including public holidays, severe weather, and academic calendars. Such events induce systematic and often asymmetric spatial shifts, such as demand surges in residential areas during holidays and demand drops in business districts, which purely data-driven backbones cannot anticipate. This results in persistent forecasting bias during event periods. Directly injecting event features into the backbone may interfere with well-learned regular patterns, whereas retraining the entire model for each event type is operationally infeasible. Recent LLM-based time-series methods~\cite{jin2310time,liu2025semantic,kim2026ctrl} remain insufficient for this setting because they either rely on costly online LLM inference or fail to explicitly exploit exogenous event semantics.

To address these challenges, we propose GLIDE, a production-ready \textbf{G}raph-\textbf{L}atent \textbf{I}nductive \textbf{D}ynamics prediction framework with \textbf{E}vent awareness. GLIDE models evolving logistics graphs in a fixed-capacity latent space while adapting to heterogeneous demand regimes and event-driven distribution shifts. Specifically, we first introduce an Inductive Graph Quantization mechanism that learns a sparse assignment matrix to dynamically project arbitrary-size physical graphs into a semantically stable Latent Token Space. Temporal dynamics are then modeled over these latent tokens, enabling inductive generalization to newly emerging SEHs while decoupling the computational cost from the number of physical nodes. Second, we design a Volume-Adaptive Mixture-of-Experts (MoE) Head that routes representations to specialized experts according to coarse-grained volume regimes, thereby alleviating the regression bias caused by zero-inflated and long-tailed demand. Finally, we develop an Event-Aware Modulation module guided by an offline LLM-constructed event knowledge bank. The module performs lightweight multiplicative corrections on a frozen forecasting backbone, reducing systematic event-period bias without costly online LLM inference or model retraining. Our main contributions are summarized as follows:

\begin{itemize}[leftmargin=1.5em,itemsep=0pt,parsep=0.2em,topsep=0.0em,partopsep=0.0em]
\item We formulate the problem as inductive learning over evolving logistics graphs and propose Graph-Guided Latent Aggregation for scalable modeling of variable-size and unseen graphs.

\item We design a Volume-Adaptive MoE Head to mitigate long-tail regression bias across heterogeneous demand regimes.

\item We introduce Event-Aware Modulation, using an offline LLM-constructed knowledge bank to correct event-period bias without online LLM inference or model retraining.

\item Extensive experiments and deployment at Meituan demonstrate superior accuracy and efficiency over SOTA methods.

\end{itemize}

\vspace{-5pt}
\section{Preliminaries}
\subsection{Concept Definitions}
\noindent\textit{\textbf{Flow Unit (FU).}}
We define an Flow Unit (FU) as a directed logistics vector connecting a specific Pick-up AOI to a Delivery AOI that exhibits valid logistics activity within a specific cyclic time window $\Delta \tau$. Unlike traditional road sensors, an FU's identity is tied to its logistics role. We denote the global set of all possible FUs as $\mathcal{V}_{global}$. In any specific inference or training instance, we operate on a subset $\mathcal{V}_S \subset \mathcal{V}_{global}$, where the composition of $\mathcal{V}_S$ is determined by the specific business query or delivery cluster.

\noindent\textit{\textbf{Scale-Effect Hotspots(SEH).}}
We organize geographically proximate and pooling-compatible FUs into scale-effect hotspots (SEHs) \cite{liang2024harvesting}. An SEH is a local flow subnetwork in which recurrent pick-up and delivery structures create substantial potential for consolidating multiple orders without excessive detours. Accordingly, the prediction target in this work is the order volume of each SEH in a 30-minute planning interval, rather than the demand of an isolated FU or a static administrative region.

\vspace{-10pt}
\subsection{Problem Formulation}
We formulate the SEHs demand forecasting as an Inductive Graph Learning problem. Given a sampled subgraph $\mathcal{G}$ containing $N_{sub}$ nodes, we observe a historical feature tensor $\mathbf{X} \in \mathbb{R}^{N_{sub} \times T_{h} \times C}$, where $T_h$ is the look-back window and $C$ includes normalized order volume and temporal markers. The objective is to learn a parameterized function $f_\theta$ that predicts future volumes $\hat{\mathbf{Y}} = f_\theta \left( \mathcal{G}, \mathbf{X}; \Phi \right)$ for the subsequent $T_{f}$ time steps, where $\Phi$ represents a set of learnable structural archetypes. A core challenge in the Meituan production environment is Zero-Shot Generalization, as the model must perform inference on a test graph $\mathcal{G}_{test} = (\mathcal{V}_{test}, \mathcal{E}_{test})$ where the nodes $\mathcal{V}_{test}$ may never have appeared during the training phase.

\section{Methodology}
\begin{figure*}
    \centering
    \includegraphics[width=0.95\linewidth]{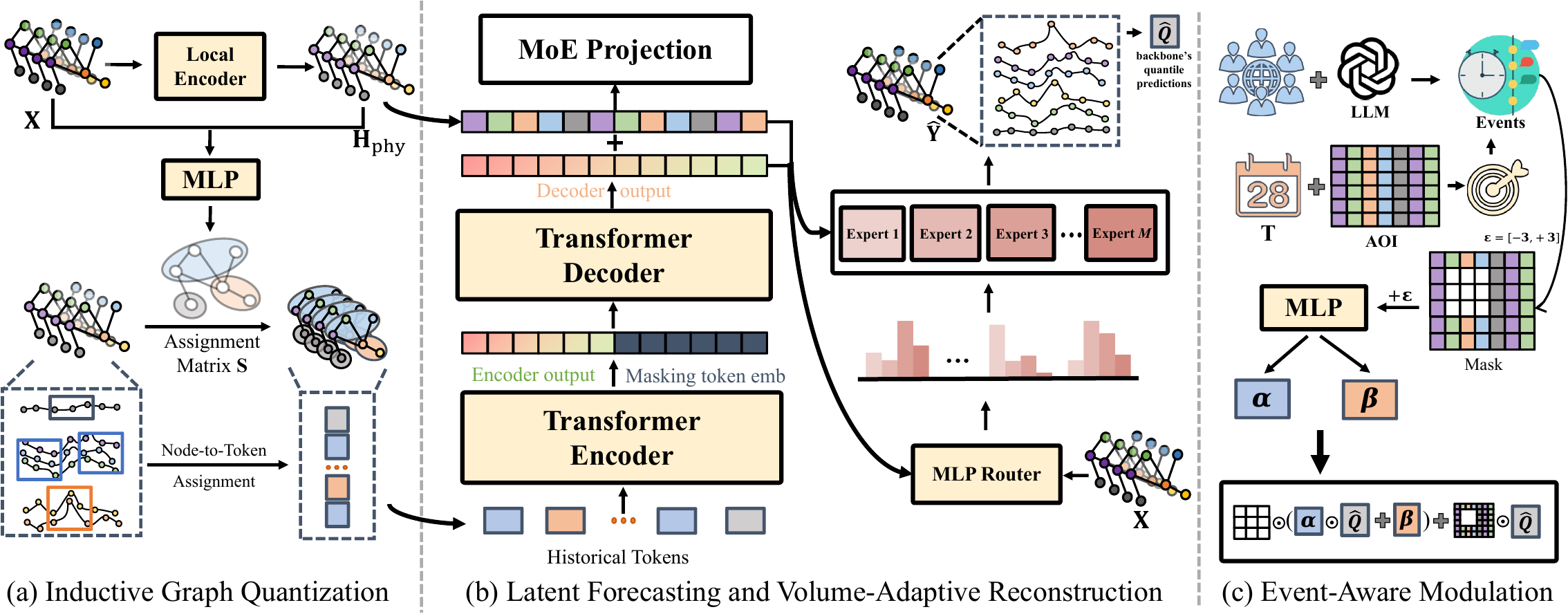}
    \vspace{-10pt}
    \caption{The Framework of GLIDE (Graph-Latent Inductive Dynamics prediction with Event awarenes).}
    \label{fig:framework}
    \vspace{-10pt}
\end{figure*}

Figure~\ref{fig:framework} presents the architecture of GLIDE, which addresses evolving topologies, scale--volume heterogeneity, and event-driven shifts. It consists of three components: (i) \textbf{Graph-Guided Latent Aggregation}, which maps variable-size graphs into fixed latent tokens; (ii) \textbf{Latent Forecasting and Volume-Adaptive Reconstruction}, which models temporal dynamics with a masked Transformer and generates SEH-level predictions through a Mixture-of-Experts head; and (iii) \textbf{Event-Aware Modulation}, which incorporates offline LLM-derived event knowledge through lightweight corrections.

\subsection{Inductive Graph Quantization}
\label{sec:quantization}

In dynamic logistics networks, the set of feasible SEHs evolves continuously, leading to frequent changes in the induced logistics subgraph $\mathcal{G} = (\mathcal{V}_S, \mathcal{E})$. In practical deployments, the size of $\mathcal{V}_S$ can range from only a few nodes to tens of thousands within a single decision window. Standard transductive GNNs are ill-suited to this setting, as they rely on fixed graph Laplacians and cannot generalize across dynamically instantiated subgraphs. Meanwhile, directly applying Transformer-style global attention over $N$ nodes incurs $O(N^2)$ complexity, rendering global-scale screening infeasible in real-time systems. To address both inductive generalization and computational scalability, we introduce an Inductive Graph Quantization module. Rather than operating on physical SEH nodes directly, the module maps them into a fixed set of $K$ latent tokens. This design effectively decouples the computational budget from the input graph size, enabling elastic inference across highly variable logistics subgraphs.

\noindent\textit{\textbf{Local Dynamics Encoding. }}Let $\mathbf{X} \in \mathbb{R}^{N \times T_{\text{h}} \times C}$ denote the input features over the look-back window $T_{\text{h}}$. To model the local dynamics of each SEH, we first apply a temporal convolution layer with kernel and stride sizes both equal to the patch size $p_t$, aggregating short-term temporal patterns. This is followed by a lightweight Graph Convolutional Network (GCN) to capture local neighborhood affinities within the logistics graph $\mathcal{G}$. The resulting local spatiotemporal representation is computed as:
\begin{equation}
    \mathbf{H}_{\text{phy}} = \text{GCN}\left( \text{Conv1d}(\mathbf{X}), \mathcal{G} \right) \in \mathbb{R}^{N \times L \times D},
\end{equation}
where $D$ is the embedding dimension and $L = \frac{T_{\text{h}}}{p_t}$ is the number of temporal patches.

\noindent\textit{\textbf{Node-to-Token Assignment.}}
To decouple the model structure from the graph size $N$, we learn a dynamic assignment matrix $\mathbf{S} \in \mathbb{R}^{N \times K}$. Unlike static clustering algorithms (e.g., METIS~\cite{karypis1998fast}) or hard discrete-quantization schemes~\cite{van2017neural}, $\mathbf{S}$ is generated dynamically based on node features and volume scales. Specifically, the Assignment Block computes the probability of assigning node $i$ to latent token $k$. To ensure distinct cluster assignments, we employ a Top-$k$ sparse masking strategy, ensuring that each SEH contributes only to its most semantically relevant latent tokens. Unlike non-differentiable hard clustering (e.g., K-Means), we employ a Sparse Attentive Assignment mechanism. This allows the model to be trained end-to-end while enforcing a strict sparsity constraint. The procedure is as follows:
\begin{equation}
\begin{aligned}
\mathbf{L}
&=\operatorname{MLP}(\mathbf{H}_{\mathrm{phy}}\oplus\mathbf{f}_{\mathrm{scale}})
\in\mathbb{R}^{N\times K},\\[-2pt]
\mathbf{S}_{i,:}
&=\operatorname{softmax}\!\left(
\operatorname{TopMask}_{r}(\mathbf{L}_{i,:})
\right).
\end{aligned}
\end{equation}

Here, $\mathbf{f}_{\text{scale}}$ represents static scale features (e.g., historical volume magnitude), allowing the model to differentiate between ``head'' and ``tail'' nodes during aggregation. $\operatorname{TopMask}_{r}(\cdot)$ retains the top-$r$ entries and masks all remaining entries with $-\infty$.

\noindent\textit{\textbf{Latent Projection.}} 
Finally, the latent representations $\mathbf{Z}_{\text{h}} \in \mathbb{R}^{K \times L \times D}$ are computed via a weighted aggregation of physical nodes. This operation acts as a Topological Bottleneck, filtering out local noise while preserving primary flow patterns:
\begin{equation}
    \mathbf{Z}_{\text{h}} = \mathbf{S}^\top \mathbf{H}_{\text{phy}},
\end{equation}
Importantly, the dimension $K$ is a hyperparameter independent of $N$, enabling the model to handle subgraphs of arbitrary sizes while adhering to a fixed computational budget. In our implementation, we set $K$ to 100. The resulting latent representation $\mathbf{Z}_{\text{h}}$ is reshaped into a tensor of size $(K \times L, D)$ and subsequently passed through the transformer model for further processing.

\subsection{Latent Forecasting and Volume-Adaptive Reconstruction}
To model the complex interactions among logistics regions, we utilize a Transformer architecture after projecting the data into latent tokens. By evolving states in the dense latent space, we can capture high-level semantic interactions (e.g., ``Business District'' $\leftrightarrow$ ``Residential Area'') more effectively than node-level interactions. The Transformer model is composed of a series of stacked layers, each consisting of a multi-head self-attention mechanism and a feed-forward neural network. 

\noindent\textit{\textbf{Global Dynamics Encoding. }} Specifically, the input sequential data $\mathbf{Z_h}$ comprises only historical records $\mathbf{Z}_{\text{h}}$. After processing by the encoder, yielding $\mathbf{Z}_{\text{e}}$, a mask token embedding $\mathbf{E}_{\text{m}}$ is concatenated with $\mathbf{Z}_{\text{e}}$ to form the complete embedding map. The decoder then applies multi-head self-attention over the entire sequence of embeddings, enabling the model to capture fine-grained global spatiotemporal correlations and dynamics across the latent tokens. The process is formalized as follows:
\begin{equation}
\begin{aligned}
    &\mathbf{Z}_{\text{e}} = \text{Encoder}(\mathbf{Z}_{\text{h}}), \\
    &\mathbf{Z}_{\text{out}} = \text{Decoder}(\text{Concat}(\mathbf{Z}_{\text{e}},\mathbf{E}_{\text{m}})).
\end{aligned}
\end{equation}
Here, $\mathbf{Z}_{\text{e}}$ represents the output of the encoder, which is subsequently fed into the decoder. The final output, $\mathbf{Z}_{\text{out}}$, is the prediction generated by the decoder, reflecting the model's learned representation of the global dynamics.

A core challenge in SEHs forecasting is the Zero-Inflated Long-Tail Distribution. A single regression head inevitably biases towards the mean (zero), underestimating critical high-demand bursts. We propose a Volume-Adaptive Mixture-of-Experts (MoE) Head to decouple the prediction logic for different volume regimes.

\noindent\textit{\textbf{Node Restoration.}} We first project the latent states back to the physical domain using the transpose of the assignment matrix $\mathbf{S}$. To preserve fine-grained local details that might have been lost during compression, we incorporate a residual connection from the local encodings:
\begin{equation}
    \hat{\mathbf{H}}_{\text{phy}} = \mathbf{S} \cdot \mathbf{Z}_{\text{out}} + \mathbf{H}_{\text{phy}}.
\end{equation}
This residual design ensures the model benefits from both global context (from Latent Tokens) and local specificity (from $\mathbf{H}_{\text{phy}}$).

To account for the heteroscedasticity of order volume across different demand regimes, we deploy a committee of $M=6$ specialized expert networks, denoted as $\{E_m\}_{m=1}^{6}$. These experts jointly capture demand patterns ranging from zero-inflated and low-volume observations to sparse but operationally critical high-volume bursts. The six demand regimes are defined according to their relevance to order pooling and dispatch optimization and collectively cover the full distribution of the target order volume $y_t$. Due to commercial confidentiality, we do not disclose the original order-count thresholds; instead, we report the interval construction strategy and normalized bucket statistics in the appendix.

\noindent\textit{\textbf{Mixture-of-Experts Head.}}
We deploy $M$ expert networks, ${E_m}_{m=1}^M$, where each expert is specialized in a specific demand range, categorized based on business relevance into 6 intervals (Exact numerical thresholds are not disclosed due to commercial constraints). The routing decision is controlled by a Gating Network $\mathcal{G}(\cdot)$, which combines the learned embedding with statistical features:
\begin{equation}
    \mathbf{w} = \text{Softmax}\left( \text{MLP}( \hat{\mathbf{H}}_{\text{phy}} \oplus \mathbf{f}_{\text{scale}}) \right) \in \mathbb{R}^{N \times M}.
\end{equation}
The final demand forecast $\hat{\mathbf{Y}}$ is obtained by a probability-weighted aggregation of the expert outputs:
\begin{equation}
    \hat{\mathbf{Y}} = \sum_{m=1}^M w_m \cdot E_m(\hat{\mathbf{H}}_{\text{phy}}).
\end{equation}
By explicitly routing high-load nodes to specific experts, GLIDE mitigates the zero-inflation bias common in standard regression models.

We formulate two distinct optimization objectives to address the different needs of real-time dispatching and capacity planning: Sequence Regression for precise temporal forecasting and Distribution Matching for capturing statistical demand properties.

\noindent\textbf{Sequence Regression.} For accurate real-time dispatching, we aim for precise temporal alignment of demand peaks. The objective is:
\begin{equation}
    \mathcal{J}_{\text{seq}} = \mathcal{L}_{\text{reg}} + \mathcal{L}_{\text{gate}}.
\end{equation}
To address the long-tail distribution of demand, we introduce a Volume-Stratified Reweighting strategy. The loss function is defined as:
\begin{equation}
    \mathcal{L}_{\text{reg}} = \frac{1}{N \cdot T_{\text{f}}} \sum_{i \in \mathcal{V}} \sum_{t \in \mathcal{T}} \Lambda(y_{i,t}) \cdot \left( \hat{y}_{i,t} - y_{i,t} \right)^2,
\end{equation}
where $\Lambda(y_{i,t})$ applies a weight to each instance, amplifying the gradients from high-volume instances. To guide the MoE head, we supervise the gating network using ground-truth volume regimes. The loss is the cross-entropy between the predicted gating probability $\pi_{i,t}$ and the target class $c_{i,t}$:
\begin{equation}
    \mathcal{L}_{\text{gate}} = - \frac{1}{N \cdot T_{\text{f}}} \sum_{i \in \mathcal{V}} \sum_{t \in \mathcal{T}} \sum_{m=1}^{M} \mathbb{I}(c_{i,t} = m) \log(\pi_{i,t}^{(m)}).
\end{equation}

\subsection{Event-Aware Modulation}
\label{sec:event_modulation}

The components above equip GLIDE with strong regular-pattern modeling, yet they remain blind to exogenous disruptions such as public holidays, severe weather, and academic calendars, which induce systematic and often asymmetric demand shifts that never appear in the historical look-back window. A naive remedy---injecting event features into the backbone and retraining---risks catastrophic forgetting of well-learned regular patterns and is operationally infeasible given the frequent emergence of new event types. Instead, we keep the backbone frozen and introduce a lightweight \textbf{Event-Aware Modulation} module that performs targeted multiplicative corrections only on event-affected spatiotemporal positions, leaving regular-period predictions untouched.

\noindent\textit{\textbf{Event Knowledge Bank.}}
To avoid any online dependency on large language models, we offline-construct an event knowledge bank via LLM distillation. While recent works adapt LLMs to time-series forecasting through representation reprogramming~\cite{jin2310time,liu2025semantic} or mixture-of-experts adapters~\cite{gu2026fm}, these approaches still require an LLM at inference time; in contrast, our knowledge bank distills the LLM's event-impact reasoning into a static lookup table that is fully decoupled from online inference. Each event in the library $\mathcal{E}$ is described by its subtype $s$ (e.g., ``public holiday'', ``rainstorm'', ``university vacation''), a spatial scope (city-wide or point-of-interest specific), and a \emph{phase range} over calendar dates (e.g., \textit{buffer\_start / start / middle / end / buffer\_end}). For each tuple $(s, p, m)$ of subtype, phase position, and meal period, the LLM produces a 16-dimensional \emph{flat impact vector} $\mathbf{v}_{s,p,m} \in \mathbb{R}^{16}$ defined over AOI-type categories, encoding how strongly each functional zone type is affected. These vectors are stored in a lookup table $\mathcal{K}$ and are completely date-agnostic, enabling a one-time construction that is reused across cities and years.

\noindent\textit{\textbf{Event Matching and Affected Mask.}}
At inference time, given a target date $t$ and a wave's AOI set, an \textbf{EventMatcher} retrieves all events in $\mathcal{E}$ whose phase range covers $t$ and whose spatial scope intersects the wave. For each retrieved event, the matcher determines the phase position $p$ and queries $\mathcal{K}$ for $\mathbf{v}_{s,p,m}$. The per-node impact score is then computed as the dot product between the wave's AOI-type distribution $\mathbf{a}_i \in \mathbb{R}^{16}$ and the impact vector:
\begin{equation}
    e_{i,t} = \mathbf{a}_i^\top \mathbf{v}_{s,p,m}, \quad \text{for event-affected } (i,t).
\end{equation}
A binary \textbf{affected time mask} $\mathbf{M} \in \{0,1\}^{N \times T_f}$ marks the spatiotemporal positions where at least one event is active; positions with $\mathbf{M}_{i,t}=0$ receive no modulation at all.

\noindent\textit{\textbf{Multiplicative Modulation.}} To obtain the probabilistic backbone output used below, we instantiate each MoE expert with $Q$ parallel outputs corresponding to ordered quantile levels $\mathcal{T}=\{\tau_q\}_{q=1}^{Q}$, combine the expert-specific quantiles using the routing weights in Eq.~(7), and optimize them with the summed pinball loss; the resulting predictions are re-sorted along the quantile dimension to ensure monotonicity, yielding $\widehat{\mathbf{Q}}\in\mathbb{R}^{N\times T_f\times Q}$. For deterministic evaluation and online dispatch, we use the predictive median $\widehat{y}_{i,t}=\widehat{Q}_{i,t,\tau=0.5}$, while CRPS is computed from the complete set of predicted quantiles.
The modulator takes the resulting backbone quantile predictions
$\widehat{\mathbf{Q}}$ together with temporal modulator inputs derived from $\{e_{i,t}\}$, and produces a scale $\boldsymbol{\alpha}$ and a shift $\boldsymbol{\beta}$. This relates to the recent line of LLM-guided residual learning for time series~\cite{kim2026ctrl}, but differs in two respects: our corrections are \emph{multiplicative} (scaling-and-shifting the quantile curve) rather than purely additive residuals, and they are gated by the affected mask so that only event-disrupted positions are altered:
\vspace{-3pt}
\begin{equation}
    \boldsymbol{\alpha}, \boldsymbol{\beta} = \mathrm{Modulator}\big(\mathbf{X}_{\text{tmp}}, \mathbf{M}, \{e_{i,t}\}\big).
\end{equation}
The corrected quantile prediction is obtained via an affine transform applied \emph{only} on affected positions:
\begin{equation}
    \tilde{\mathbf{Q}}_{i,t} = \mathbf{M}_{i,t} \odot \big(\boldsymbol{\alpha}_{i,t} \odot \hat{\mathbf{Q}}_{i,t} + \boldsymbol{\beta}_{i,t}\big) + \big(1-\mathbf{M}_{i,t}\big) \odot \hat{\mathbf{Q}}_{i,t},
\end{equation}
followed by a re-sorting step to preserve quantile monotonicity. This design guarantees two properties: (i) \textbf{targeted correction}---only event-affected positions are altered, so regular-period forecasts inherit the backbone's accuracy verbatim; and (ii) \textbf{frozen-backbone compatibility}---since the backbone parameters are untouched, the modulator can be trained or updated independently without risking the learned logistics archetypes.

\section{Experiments}
\subsection{Experimental Setup}

\noindent\textbf{Datasets.} We evaluate GLIDE on six large-scale, real-world datasets collected from the Meituan platform, covering the period from July 1 2025 to July 1, 2026. We distinguish between \textit{Weekday} (Wd) and \textit{Weekend} (We) to account for heteroscedasticity in order distributions. Unlike static sensor benchmarks like PEMS-BAY, our data consists of {dynamic logistics subgraphs} where nodes emerge and vanish. Table~\ref{tab:datasets} details the structural heterogeneity, specifically the wide range of $|V_S|$, which necessitates the inductive capacity of GLIDE. To enhance the model’s zero-shot generalization and robustness to density shifts, we implement two sampling strategies. Specifically, in Uniform Structural Sampling, we randomly sample $\eta \in \{20\%, 40\%, 60\%, 80\%\}$ of nodes from each subgraph to simulate partial observation scenarios. In Volume-Prioritized Sampling, we select the top $\eta$ nodes by volume to assess the model's ability to prioritize `Head' SEHs critical for pooling efficiency.

\begin{table}[t!]
\centering
\caption{Statistics of the Meituan OFD Datasets. The extreme variance in $|V_S|$ (e.g., 1 to 5,915) highlights the challenge for transductive baselines.}
\vspace{-10pt}
\label{tab:datasets}
\small
\resizebox{0.8\columnwidth}{!}{
\begin{tabular}{lccc}
\toprule 
\textbf{City ID} & \textbf{Type} & \textbf{\# Subgraphs} & \textbf{Nodes per Subgraph $|V_S|$} \\ 
\midrule
City A & Wd / We  & 64 & 1 -- 5,915 \\ 
City B & Wd / We  & 79 & 1 -- 3,224 \\ 
City C & Wd / We  & 18 & 6 -- 3,934 \\ 
\bottomrule
\end{tabular}
}
\vspace{-10pt}
\end{table}

\noindent\textbf{Baselines.} We compare GLIDE against five categories of SOTA models: (i) \textit{Statistical}: \textbf{HA}. (ii) \textit{MLP-based}: \textbf{STID}~\cite{shao2022spatial} and \textbf{DLinear}~\cite{zeng2023transformers}. (iii) \textit{Transformer-based}: \textbf{Informer}~\cite{zhou2021informer}, \textbf{Autoformer}~\cite{wu2021autoformer}, and \textbf{iTransformer}~\cite{liu2024itransformer}. (iv) \textit{ST-Graph}: \textbf{STGCN}~\cite{yu2017spatio} and \textbf{DSTSGNN}~\cite{lan2022dstagnn}.(\textit{Note: we train separate instances for each subgraph due to their fixed-graph constraints.}) (v) \textit{Inductive Foundational}: \textbf{UniFlow}~\cite{yuan2025uniflowfoundationmodelunified}, which supports cross-graph training, and is trained using the same parameter setup as GLIDE for a fair comparison.

\noindent\textbf{Implementation Details.} GLIDE uses a 2-layer Encoder/Decoder ($D=64$). We use Adam ($LR=10^{-4}$, weight decay $10^{-5}$) with 200 epochs and early stopping (patience=10). Baselines are grid-searched across $LR \in \{10^{-2}, 10^{-3}, 10^{-4}\}$. All experiments ran on NVIDIA Tesla V100 GPUs (32GB). The event knowledge bank is constructed offline via LLM distillation (GPT-4o-mini), covering 123 events across five categories: public holidays, campus calendar events, weather events, other social events, and platform promotions. Details are provided in Appendix~\ref{sec:event-setup-detail}. Training and inference efficiency are reported in Appendix~\ref{sec:efficiency}, with the full training-cost comparison in Appendix~\ref{sec:train_cost}.

\begin{table*}[t!]
\centering

\begin{minipage}[t]{0.725\textwidth}
\vspace{0pt}
\centering
\captionof{table}{
Overall performance on multi-city datasets.
We report Mean Absolute Error (MAE) and Root Mean Squared Error (RMSE).
Lower values indicate better performance.
\textbf{Bold} and \underline{underlined} values denote the best and
second-best results, respectively.
}
\vspace{-8pt}
\label{tbl:prediction_results}

\resizebox{\linewidth}{!}{
\begin{tabular}{l|cc|cc|cc|cc|cc|cc}
\toprule
\multirow{3}{*}{\textbf{Model}}
& \multicolumn{6}{c|}{\textbf{Weekday Scenarios (Stable)}}
& \multicolumn{6}{c}{\textbf{Weekend Scenarios (Bursty)}} \\
\cmidrule(lr){2-7}
\cmidrule(lr){8-13}

& \multicolumn{2}{c|}{\textbf{City A-Wd}}
& \multicolumn{2}{c|}{\textbf{City B-Wd}}
& \multicolumn{2}{c|}{\textbf{City C-Wd}}
& \multicolumn{2}{c|}{\textbf{City A-We}}
& \multicolumn{2}{c|}{\textbf{City B-We}}
& \multicolumn{2}{c}{\textbf{City C-We}} \\

& MAE & RMSE
& MAE & RMSE
& MAE & RMSE
& MAE & RMSE
& MAE & RMSE
& MAE & RMSE \\
\midrule

\textbf{HA}
& 1.335 & 1.895
& 1.246 & 1.781
& 1.186 & 1.689
& 1.455 & 2.091
& 1.293 & 1.892
& 1.182 & 1.649 \\

\textbf{Autoformer}
& 1.300 & 1.967
& 1.193 & 1.970
& 1.280 & 1.834
& 1.371 & 2.100
& 1.299 & 2.256
& 1.141 & 1.639 \\

\textbf{iTransformer}
& 1.166 & 1.751
& 1.160 & 1.712
& 1.106 & 1.617
& \underline{1.294} & 1.956
& 1.200 & 1.991
& 1.102 & 1.619 \\

\textbf{Informer}
& 1.273 & 1.965
& 1.190 & 1.952
& 1.222 & 1.960
& 1.351 & 2.042
& 1.259 & 2.186
& 1.126 & 1.641 \\

\textbf{DLinear}
& 1.162 & 1.736
& \underline{1.112} & \underline{1.700}
& 1.014 & 1.590
& 1.295 & 1.947
& 1.160 & 1.873
& 0.995 & 1.511 \\

\textbf{STID}
& \underline{1.147} & \underline{1.729}
& 1.162 & 1.736
& 0.997 & 1.581
& 1.297 & 1.946
& \underline{1.139} & \underline{1.849}
& \textbf{0.967} & \textbf{1.501} \\

\textbf{STGCN}
& 1.316 & 1.899
& 1.167 & 1.955
& 1.037 & 1.684
& 1.430 & 2.099
& 1.225 & 2.181
& 0.996 & 1.615 \\

\textbf{DSTSGNN}
& 1.257 & 1.891
& 1.149 & 1.878
& 1.020 & 1.622
& 1.372 & 2.069
& 1.221 & 2.074
& 1.004 & 1.591 \\

\textbf{UniFlow}
& 1.166 & 1.757
& 1.161 & 1.901
& \underline{0.996} & \underline{1.580}
& 1.297 & \underline{1.944}
& 1.147 & 1.900
& 1.001 & 1.601 \\

\midrule
\rowcolor{gray!15}
\textbf{GLIDE (Ours)}
& \textbf{1.131} & \textbf{1.691}
& \textbf{1.055} & \textbf{1.624}
& \textbf{0.973} & \textbf{1.504}
& \textbf{1.284} & \textbf{1.939}
& \textbf{1.132} & \textbf{1.819}
& \underline{0.995} & \underline{1.506} \\
\bottomrule
\end{tabular}
}
\vspace{-10pt}
\end{minipage}
\hfill
\begin{minipage}[t]{0.255\textwidth}
\vspace{0pt}
\centering

\captionof{table}{
Representative zero-shot transfer from City A to unseen Cities B
and C. Each entry reports MAE/RMSE. 
}
\vspace{-10pt}
\label{tab:zero_shot}

\scriptsize
\setlength{\tabcolsep}{3.0pt}
\renewcommand{\arraystretch}{1.08}

\resizebox{\linewidth}{!}{
\begin{tabular}{l|c|c}
\toprule
\textbf{Model}
& \textbf{Target B}
& \textbf{Target C} \\
\midrule

\multicolumn{3}{c}{
\textbf{Weekday Scenarios (Stable)}
} \\
\midrule

HA
& 1.246/1.781
& 1.186/1.689 \\

UniFlow-ZS
& 1.221/1.777
& 1.196/1.928 \\

\rowcolor{gray!15}
\textbf{GLIDE-ZS}
& \textbf{1.048/1.618}
& \textbf{1.022/1.665} \\

\textit{GLIDE-Sup.}
& \textit{1.055/1.624}
& \textit{0.973/1.504} \\

\midrule

\multicolumn{3}{c}{
\textbf{Weekend Scenarios (Bursty)}
} \\
\midrule

HA
& 1.293/1.892
& 1.182/1.649 \\

UniFlow-ZS
& 1.244/1.935
& 1.219/1.732 \\

\rowcolor{gray!15}
\textbf{GLIDE-ZS}
& \textbf{1.130/1.788}
& \textbf{1.130/1.788} \\

\textit{GLIDE-Sup.}
& \textit{1.132/1.819}
& \textit{0.995/1.506} \\

\bottomrule
\end{tabular}
}

\vspace{2pt}

\end{minipage}
\end{table*}

\subsection{Overall Performance Comparison}

\textit{\textbf{Task Specification.}}
To comprehensively evaluate the model's robustness across distinct operational cycles, we test our model in two tasks that mirror real-world dispatching requirements:
\begin{itemize}[leftmargin=*]
    \item \textbf{Weekday Routine (Stable Dynamics):} For weekday datasets, which exhibit repetitive commuting and office-hour patterns, we utilize a longer horizon strategy. We use a historical window of $T_h=20$ days to forecast the subsequent $T_f=5$ days. This tests the model's ability to capture long-range periodicity and sustained demand trends.
    \item \textbf{Weekend Lifestyle (Bursty Dynamics):} Weekend patterns are characterized by high entropy and impulsive ordering behaviors. We adopt a responsive short-horizon strategy, using $T_h=8$ days to predict the next $T_f=4$ days. This setting stresses the model's responsiveness to rapid distribution shifts and irregular peaks.
\end{itemize}

We further analyze how the length and temporal ordering of the input window affect the cost and accuracy trade-off in Appendix~\ref{sec:reprogram}.
\noindent\textit{\textbf{Results Analysis.}}
Table \ref{tbl:prediction_results} compares the performance of GLIDE against state-of-the-art baselines across heterogeneous urban environments. GLIDE achieves the best performance in 10 out of 12 metrics. The results highlight three specific architectural advantages:\textbf{(i)Inductive Structure Learning vs. Transductive Overfitting.} Traditional GNNs (STGCN, DSTSGNN) exhibit suboptimal performance, particularly in City A (RMSE 1.899 vs. GLIDE 1.691). This confirms the limitation of the transductive setting: because these models must be trained separately for each disjoint subgraph, they suffer from data fragmentation and cannot exploit city-wide patterns. In contrast, GLIDE's \textit{Latent Inductive Graph} mechanism allows it to learn a shared set of ``Logistics Archetypes'' across all subgraphs, effectively transferring knowledge from high-density regions to sparse ones. \textbf{(ii) The "Simplicity vs. Complexity" Paradox.} 
Consistent with recent findings in time series forecasting, simple linear baselines (DLinear, STID) outperform complex Transformer-based models (Informer, Autoformer) on highly dynamic data. This is attributed to the low signal-to-noise ratio in ``Tail'' SEHs, where heavy attention mechanisms tend to overfit stochastic noise. GLIDE overcomes this by projecting noisy physical nodes into a robust latent token space before applying temporal evolution. This acts as a regularization bottleneck, filtering out node-level noise while preserving macro-level dynamics, allowing GLIDE to outperform DLinear by a significant margin in high-complexity scenarios. \textbf{(iii) Scalability and Robustness.} We observe that lightweight baselines like STID remain competitive on smaller, simpler graphs (e.g., City C-We). However, as network complexity and heterogeneity increase (City A and B), the performance gap widens in GLIDE's favor. This validates that our \textit{Volume-Adaptive MoE} and \textit{Graph Quantization} are particularly effective for large-scale, heterogeneous networks where static linear mappings fail to capture the evolving structural dependencies.

\noindent\textit{\textbf{Zero-Shot Capabilities.}} To evaluate GLIDE's capability to learn universal logistics dynamics independent of specific city topologies, we conducted extensive Zero-Shot Transfer experiments. As shown in Table \ref{tab:zero_shot}, we trained the model on a source city and directly applied it to two unseen target cities. GLIDE (Zero-shot) significantly outperforms the baseline foundation model (UniFlow~\cite{yuan2025uniflowfoundationmodelunified}) across all transfer pairs. Remarkably, the performance gap between the zero-shot model and the fully supervised upper bound is minimal.
This confirms that our Graph-Guided Latent Aggregation successfully decouples structural dependencies from temporal dynamics, learning a set of universal logistics archetypes (e.g., residential morning peaks, commercial lunch bursts) that are invariant across different areas. While UniFlow is designed for cross-graph generalization, it struggles to match GLIDE's precision in the zero-shot setting (e.g., UniFlow MAE 1.221 vs. GLIDE 1.048 on A$\rightarrow$B). This suggests that generic time-series foundation models, which often ignore the explicit logistics graph topology, fail to capture the spatial constraints inherent in food delivery networks. GLIDE's design, which explicitly quantizes the graph into semantically meaningful tokens, proves to be a more effective inductive bias for transfer learning in physical systems. Additionally, we observe a distinct asymmetry in transfer performance. Transferring from a larger, more complex dataset (City A, 93k nodes) to a smaller one (City B, 34k nodes) yields exceptional results. In Panel A, the zero-shot model on City B-Weekday (MAE 1.048) even slightly surpasses the model trained on City B itself (MAE 1.055). We attribute this to the richer diversity of flow patterns in City A, which allows the model to learn a more robust manifold of latent tokens, effectively acting as a form of data augmentation for the smaller target city. Additional transfer pairs using City~B and~C as source domains are provided in Appendix~\ref{sec:zero_shot_extra}.

\vspace{-5pt}
\subsection{Ablation Study on Model Design}
\begin{figure}
    \centering
    \includegraphics[width=1\linewidth]{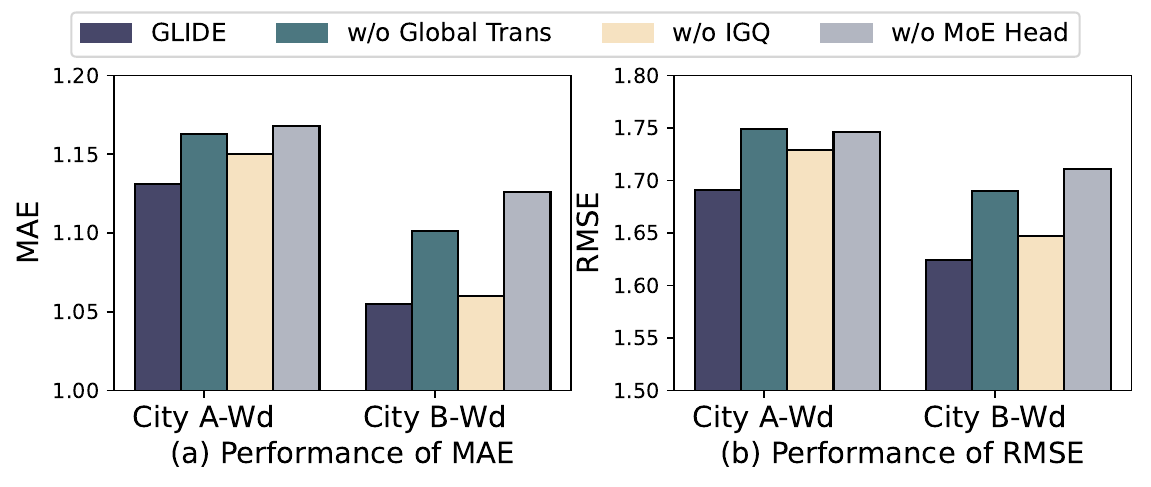}
    \vspace{-10pt}
    \caption{Performance comparison of the GLIDE architecture and its ablation variants across City A-Wd and City B-Wd.}
    \label{fig:ablation}
    \vspace{-10pt}
\end{figure}

To verify the effectiveness of the core architectural components in GLIDE, we conduct an ablation study by systematically removing or replacing key modules. We construct the following three variants: \textbf{(i) w/o Inductive Graph Quantization (w/o IGQ):} We replace the dynamic node-to-token assignment module with the static METIS graph partitioning algorithm. This variant tests the importance of learning dynamic, semantic clusterings versus using fixed topological partitions. \textbf{(ii) w/o Global Transformer (w/o Trans):} We remove the Transformer layer operating in the latent space. The model relies solely on the local GCN and temporal convolutions for inference. This evaluates the necessity of modeling global long-range dependencies. \textbf{(iii) w/o Mixture-of-Experts (w/o MoE):} We replace the Volume-Adaptive MoE head with a standard MLP regression head. This validates the MoE's ability to handle zero-inflated and long-tail demand distributions. The experimental results are presented in Figure~\ref{fig:ablation}.

As shown in Figure \ref{fig:ablation}, the full GLIDE architecture consistently outperforms all ablation variants, confirming that each component is essential for robust forecasting in complex logistics environments. Notably, the most severe performance degradation is observed in the \textbf{w/o MoE Head} variant, where the MAE notably increases across all datasets (e.g., error surge in City B-Wd); this sharp decline empirically validates that a standard shared MLP is ill-equipped to handle the heterogeneous zero-inflated distribution, as it inevitably biases predictions towards the mean, whereas our volume-adaptive routing successfully disentangles the learning of head and tail regimes. The removal of the global temporal modeling module (\textbf{w/o Global Trans}) also leads to significant error increases (e.g., rising MAE in City A-Wd), demonstrating that local operations alone cannot capture long-range city-wide dependencies, such as the ripple effects of demand propagating between non-adjacent functional zones. Furthermore, the comparison with \textbf{w/o IGQ} reveals that our dynamic quantization is superior to static topological partitioning (METIS), as the static approach relies on rigid physical boundaries that sever correlations between geographically distant but functionally synchronized areas (e.g., distributed commercial centers), leading to suboptimal latent representations. Ultimately, the synergy of these components allows GLIDE to simultaneously address the challenges of distribution heteroscedasticity, global temporal evolution, and inductive structural generalization. More volume-adaptive routing results can refer to the appendix: a volume-bucket error analysis is provided in Appendix~\ref{sec:tail_robustness}, and a router-visualization case study (Appendix~\ref{sec:case_study}). 

\vspace{-5pt}
\subsection{Event-Scenario Evaluation}
\label{sec:event_eval}

Beyond the regular-pattern benchmarks above, we evaluate the Event-Aware Modulation module on event-disrupted periods, where the frozen backbone alone is expected to be systematically biased. We construct backtests on City~A covering five event categories: \textbf{public holidays} (New Year, Qingming, Labor Day, Dragon Boat), \textbf{weather events} (rainstorms, sustained heat), \textbf{platform promotions} (Double 11/12), \textbf{social events} (Christmas Eve, Qixi Festival), and \textbf{overlapping event days} (periods where World Cup and university vacation overlap). Note that the Dragon Boat label period (June~22--26) also includes a rainstorm event on June~22, so its results reflect the combined effect of holiday and weather modulation. Following the affected-time-step protocol, we restrict evaluation to data samples where at least one event is active ($\mathbf{M}_{i,t}=1$), which isolates modulator's contribution from regular-period predictions.

\begin{table}[t!]
\centering
\caption{Event-Scenario Evaluation on City A across diverse event types (affected timesteps). We compare History (quantile prior), GLIDE backbone (Phase-1, P1), and GLIDE+EventModulator (Phase-2, P2). $\uparrow$ denotes relative CRPS improvement of P2 over P1; positive means P2 is better.}
\vspace{-10pt}
\label{tab:event_main}
\setlength{\tabcolsep}{2pt} %
\resizebox{1\columnwidth}{!}{
\begin{tabular}{l|cc|c|ccc}
\toprule
\textbf{Event Type} & \textbf{Event} & \textbf{ATM\%} & \textbf{N} & \textbf{Hist CRPS} & \textbf{P1 CRPS} & \textbf{P2 CRPS} ($\uparrow$) \\
\midrule
\multicolumn{7}{l}{\textit{Public Holidays (City-wide)}} \\
Holiday & New Year    & 55.9\% & 640,927   & 1.3705 & 1.2946 & \textbf{1.2702} (+1.9\%) \\
Holiday & Qingming    & 40.4\% & 625,165   & 1.1878 & 1.1243 & \textbf{1.0685} (+5.0\%) \\
Holiday & Labor Day   & 66.0\% & 1,024,930 & 1.1407 & 1.0771 & \textbf{0.9990} (+7.3\%) \\
Holiday & Dragon Boat & 100\%  & 647,190   & 1.6368 & 1.5435 & \textbf{1.5227} (+1.3\%) \\
\midrule
\multicolumn{7}{l}{\textit{Weather Events (City-wide)}} \\
Weather & Rainstorm + Heat & 100\% & 461,830 & 1.6021 & 1.5518 & \textbf{1.5373} (+0.9\%) \\
Weather & Sustained Heat   & 100\% & 498,035 & 1.6566 & 1.5981 & \textbf{1.5722} (+1.6\%) \\
\midrule
\multicolumn{7}{l}{\textit{Platform Promotions (City-wide)}} \\
Promo  & Double 11       & 80\% & 410,360 & 1.5268 & 1.4818 & \textbf{1.4671} (+1.0\%) \\
Promo  & Double 12       & 60\% & 400,925 & 1.5183 & 1.4698 & \textbf{1.4559} (+0.9\%) \\
\midrule
\multicolumn{7}{l}{\textit{Social Events (City-wide)}} \\
Social & Christmas Eve  & 80\% & 394,520 & 1.5218 & 1.4771 & \textbf{1.4705} (+0.4\%) \\
Social & Qixi Festival  & 78\% & 813,060 & 1.7039 & 1.6275 & 1.6385 (-0.7\%) \\
\midrule
\multicolumn{7}{l}{\textit{Overlapping Event Days (World Cup + University Vacation)}} \\
Overlap & Period 1 (Jun 8--12)  & 60\%  & 369,225 & 1.4411 & 1.3993 & \textbf{1.3960} (+0.2\%) \\
Overlap & Period 2 (Jun 15--19) & 100\% & 373,530 & 1.6180 & 1.5581 & \textbf{1.5281} (+1.9\%) \\
Overlap & Period 3 (Jun 29--Jul 3) & 100\% & 390,490 & 1.5465 & 1.5139 & \textbf{1.5124} (+0.1\%) \\
\bottomrule
\end{tabular}
}
\vspace{-10pt}
\end{table}

\noindent\textit{\textbf{Three-Way Comparison.}}
Table~\ref{tab:event_main} compares three configurations: \textit{History} (quantile prior from historical observations), \textit{GLIDE} (the Phase-1 backbone), and \textit{GLIDE+Event} (Phase-2 with the Event-Aware Modulator) across all five event categories. \textbf{(i) Consistent improvement.} GLIDE+Event outperforms the backbone on CRPS across all event types except Qixi Festival ($-0.7\%$), with larger gains on events that exhibit stronger disruption (e.g., Labor Day $+7.3\%$, Qingming $+5.0\%$, sustained heat $+1.6\%$). Notably, the modulator generalizes beyond holidays to \emph{weather} ($+0.9\%$ to $+1.6\%$) and \emph{platform promotions} ($+0.9\%$ to $+1.0\%$), confirming that the LLM-distilled event knowledge bank captures impact patterns across diverse event semantics, not just calendar holidays. \textbf{(ii) Overlapping event days.} During periods where multiple events co-occur (World Cup + university vacation), the modulator yields $+0.1\%$ to $+1.9\%$ gains, demonstrating robustness under compound event scenarios. \textbf{(iii) Qixi anomaly.} The only negative case is Qixi Festival ($-0.7\%$), likely because its impact vector in the knowledge bank overestimates the demand shift; this highlights the sensitivity of the modulator to the accuracy of the LLM-generated impact vectors. \textbf{(iv) Cross-city generalization.} When the City~A-trained backbone and modulator are transferred zero-shot to City~B and City~C, the modulator still yields positive CRPS improvements across nearly all event types (e.g., City~B Labor Day $+10.1\%$, City~C Qingming $+5.6\%$, weather events $+0.8\%$ to $+1.1\%$), confirming that the event-knowledge-bank paradigm generalizes across cities with different topologies. \textbf{(v) Event library ablation.} Replacing the enriched 123-event library with a legacy 27-event library degrades performance: on Dragon Boat, the CRPS improvement flips from $+1.3\%$ to $-0.4\%$, because newly added weather events (e.g., a rainstorm on June~22) raise coverage from 32.8\% to 100\%, directly demonstrating that event-type expansion makes more samples benefit from modulation. More results are provided in Appendix~\ref{sec:cross_city} and~\ref{sec:event_ablation}.

\vspace{-5pt}
\section{Online Deployment and Evaluation}

\noindent\textit{\textbf{Deployment Setup}}
Beyond offline benchmarking, we deployed GLIDE on the production dispatching system of Meituan. The online system operates on a two-timescale architecture~\cite{liang2024harvesting}: every 30 minutes, GLIDE forecasts SEH-level demand to construct workload-balanced solver units via spatial decomposition; every 30 seconds, the dispatch system solves the many-to-one assignment (MOA) problem within each unit. The scale of these dynamically generated subgraphs is highly constrained, typically containing fewer than 300 nodes (detailed data statistics are presented in Figure~\ref{fig:online_dist}(b) in the appendix~\ref{sec:online_data}). We compare GLIDE against the Incumbent Industrial Baseline, a robust ensemble model currently deployed for traffic forecasting. We evaluate two dimensions: \textbf{(i) Forecasting Accuracy} (the upstream prerequisite for balanced partitioning) and \textbf{(ii) Downstream Dispatch Impact} (the end-to-end MOA improvement driven by accurate forecasts and balanced decomposition).

\begin{table}[t!]
\centering
\caption{Online Regression Performance. We report MAE, RMSE, and MAPE. The ``Critical Interval'' represents the AFUs most sensitive to dispatch optimization.}
\vspace{-10pt}
\label{tab:online_regression}
\small 

\resizebox{0.48\textwidth}{!}{\begin{tabular}{ll|ccc|ccc}
\toprule
& & \multicolumn{3}{c|}{\textbf{Weekday Scenario}} & \multicolumn{3}{c}{\textbf{Weekend Scenario}} \\
\textbf{Range} & \textbf{Model} & \textbf{MAE} & \textbf{RMSE} & \textbf{MAPE} & \textbf{MAE} & \textbf{RMSE} & \textbf{MAPE} \\ 
\midrule
\multirow{3}{*}{\textbf{$[0, \infty)$}} & Baseline & 2.354 & 3.759 & 0.493 & 2.522 & \textbf{3.715} & 0.484 \\
& \textbf{GLIDE} & \textbf{2.191} & \textbf{3.478} & \textbf{0.445} & \textbf{2.504} & 3.770 & \textbf{0.444} \\ 
& \textit{Improv. (\%)} & \textit{+6.92\%} & \textit{+7.48\%} & \textit{+9.74\%} & \textit{+0.71\%} & \textit{-1.48\%} & \textit{+8.26\%} \\
\midrule
\multirow{3}{*}{\textbf{\makecell{Critical\\ Interval}}} & Baseline & 3.465 & 4.884 & 0.404 & 3.507 & 4.683 & 0.353 \\
& \textbf{GLIDE} & \textbf{3.139} & \textbf{4.225} & \textbf{0.354} & \textbf{3.465} & \textbf{4.593} & \textbf{0.333} \\ 
& \textit{Improv. (\%)} & \textit{+9.41\%} & \textit{+13.49\%} & \textit{+12.38\%} & \textit{+1.20\%} & \textit{+1.92\%} & \textit{+5.67\%} \\
\bottomrule
\end{tabular}
}
\vspace{-19pt}
\end{table}

\noindent\textit{\textbf{Forecasting Regression Results}} Table~\ref{tab:online_regression} reports the deterministic forecasting accuracy. In the weekday scenario, GLIDE significantly outperforms the baseline across all metrics. Notably, in the critical interval (the medium-to-high demand regime most sensitive to dispatch optimization), we achieve a 12.38\% reduction in MAPE ($0.404 \to 0.354$) compared to the baseline. In the weekend scenario, GLIDE achieves a lower MAPE ($0.444$ vs.\ $0.484$), indicating more robust relative accuracy for route planning. These improvements confirm that accurate SEH-level demand forecasts provide the foundation for balanced solver partitioning. Distribution-level metrics (Coverage Rate, CRPS, etc.) are provided in Appendix~\ref{sec:online_distribution}.

\vspace{-10pt}
\begin{table}[htbp]
\centering
\caption{Bundle-size distribution in the online A/B test. Base and Test denote the incumbent and GLIDE-guided dispatch strategies, respectively; $\Delta$ is Test$-$Base in percentage points.}
\label{tab:dispatch_impact}
\vspace{-6pt}
\small
{
\setlength{\tabcolsep}{2.5pt}
\renewcommand{\arraystretch}{0.95}
\begin{tabular}{@{}lccccccc@{}}
\toprule
\textbf{Bundle size} & \textbf{1} & \textbf{2} & \textbf{3} &
\textbf{4} & \textbf{5} & \textbf{6} & \textbf{7} \\
\midrule
Base (\%)  & 46.42 & 37.95 & 11.16 & 3.76 & 0.71 & 0.00 & 0.00 \\
Test (\%)  & 13.96 & 39.32 & 26.07 & 13.87 & 4.34 & 1.48 & 0.96 \\
$\Delta$ (pp) & $-32.46$ & $+1.37$ & $+14.91$ & $+10.11$ &
$+3.63$ & $+1.48$ & $+0.96$ \\
\bottomrule
\end{tabular}
}
\vspace{-10pt}
\end{table}

\label{sec:downstream-dispatch-impact}
\noindent\textit{\textbf{Downstream Dispatch Results.}} We further evaluated whether the forecasting improvement translated into better downstream dispatch decisions. The algorithmic framework executes at the granularity of aggregated solver units, where balanced workloads enable deeper combinatorial search within the fixed 30-second online budget. In a large-network setting without decomposition, the number of orders assignable to a single courier per dispatch round is severely limited; the aggregated assignment framework significantly expands the space for large-bundle (multi-order) matching. We conducted online A/B testing across three cities, comparing the proposed framework (GLIDE forecast $\to$ SEH-based decomposition $\to$ MOA solver) against the incumbent single-network dispatch baseline, under the same 30-second dispatch cadence and production constraints. Table~\ref{tab:dispatch_impact} reports the bundle-size distribution of aggregated-group waybills under the A/B test (\textit{base} vs.\ \textit{test}). \textbf{(i)} Compared to the base group, the fraction of single-order waybills (bundle size~$=$~1) drops sharply by $-32.46$pp, indicating that the SEH-based decomposition substantially pools single orders into multi-order bundles. \textbf{(ii)} The fractions of bundle sizes 2 through 7 all rise, with size~3 gaining the most ($+14.91$pp), followed by size~4 ($+10.11$pp) and size~5 ($+3.63$pp); the test group further produces bundles of size~6 and~7 ($+1.48$pp and $+0.96$pp) that are absent in the base group. \textbf{(iii)} Overall, the test group exhibits a higher degree of aggregation---the many-to-one ratio distribution shifts rightward toward larger bundles---confirming that SEH-level forecasting accuracy translates into operational value: more accurate workload estimates yield better-balanced, weakly coupled solver units, which in turn expose more high-quality multi-order assignments within each real-time dispatch round. The P95 dispatch latency was maintained within the operational budget, and on-time delivery quality was not compromised.

\vspace{-10pt}
\section{Related Work}

\noindent\textit{\textbf{Time Series Forecasting and Long-Tail Modeling.}} Time series forecasting has evolved from statistical methods, such as ARIMA and Historical Average (HA), to deep architectures based on RNNs, MLPs, and CNNs. More recently, Transformer-based models, including Informer~\cite{zhou2021informer}, Autoformer~\cite{wu2021autoformer}, and iTransformer, have improved long-range temporal modeling through self-attention. However, they typically model individual series independently or capture cross-series interactions implicitly, without explicitly considering the spatial topology of physical systems. Real-world demand forecasting is also challenged by long-tailed distributions dominated by zero- and low-volume observations. Conventional techniques, such as loss re-weighting~\cite{cao2019learning}, over-sampling, and representation-classifier decoupling~\cite{kang2019decoupling}, are mainly designed for classification and may be unstable for high-variance time-series regression. Although Zero-Inflated Projection distinguishes zero and non-zero targets, it does not explicitly model heterogeneous demand levels. 

\noindent\textit{\textbf{Graph-Based Spatiotemporal Forecasting.}} Spatiotemporal forecasting expands on time-series analysis by explicitly modeling spatial correlations. Early approaches employed grid-based CNNs to process city-wide crowd flows as images. Later, Graph Neural Networks (GNNs)---from spectral GCNs to attention-based variants~\cite{hamilton2017inductive,velivckovic2017graph}---became the standard for non-Euclidean data, using message passing on road networks to capture dependencies. Representative traffic forecasting models include DCRNN~\cite{li2017diffusion}, STGCN~\cite{yu2017spatio}, Graph WaveNet~\cite{wu2019graph}, AGCRN~\cite{bai2020adaptive}, attention-based ST-GCRNs~\cite{liu2023attention}, STAEformer~\cite{liu2023spatio}, and PDFormer~\cite{jiang2023pdformer}. Despite their success, these models are primarily transductive, relying on a static adjacency matrix and a fixed set of nodes. While recent Spatiotemporal Foundation Models aim to generalize across cities through large-scale pre-training, they often assume a stable sensor topology during inference. In contrast, in real-world networks, nodes continuously emerge and disappear, making fixed-graph models less effective.

\noindent\textit{\textbf{LLMs for Time Series Forecasting.}} Motivated by the success of large language models, a recent line of work adapts LLMs to time-series forecasting. Time-LLM~\cite{jin2310time} reprograms a frozen LLM with modality-specific prompts to align time-series tokens with language representations. Subsequent studies further enhance semantic alignment between numerical series and LLM priors~\cite{liu2025semantic}, introduce frequency-aware mixture-of-experts adapters~\cite{gu2026fm}, or use LLM-guided residual learning to refine a base forecaster~\cite{kim2026ctrl}. Recent work also examines low-frequency modeling with LLMs~\cite{you2026loft} and analyzes how tokenizer and backbone choices affect LLM-based forecasting~\cite{zhang2026tokenizer}.  However, these methods either require an LLM at inference time or do not explicitly encode exogenous event semantics (e.g., holidays, weather, academic calendars), which are the primary source of demand disruption in OFD.

\section{Conclusion}
In this paper, we presented GLIDE, a scalable and event-aware forecasting framework for evolving on-demand delivery networks. Graph-Guided Latent Aggregation enables scalable inductive forecasting, while the Volume-Adaptive MoE and Event-Aware Modulation address long-tail bias and event-driven shifts, respectively. Experiments on real-world datasets and online deployment in Meituan demonstrate the superior performance of GLIDE.

\newpage
\bibliographystyle{ACM-Reference-Format}
\bibliography{main}

\appendix

\section{Appendix}

\subsection{Additional Statement for SEH as the forecasting and dispatch unit}
The prediction granularity defined in the paper is determined by the downstream dispatch problem. Meituan's dispatch system executes once every 30 seconds and solves a many-to-one assignment (MOA) problem between pending orders, including feasible order combinations, and candidate couriers. Directly searching over all regional order--courier relationships produces a prohibitively large combinatorial space under the strict online latency budget. SEHs expose the local order structure that is most relevant to pooling---including flow direction, spatial compatibility, and demand concentration---and therefore provide semantically meaningful optimization atoms. Forecasting SEH-level demand allows the system to anticipate where dense pooling opportunities will emerge, configure solver workloads before each planning interval, and devote the fixed online computation budget to deeper search within promising subspaces.

\subsection{Computational Efficiency}
\label{sec:efficiency}

\begin{figure}[h]
    \centering
    \includegraphics[width=1\linewidth]{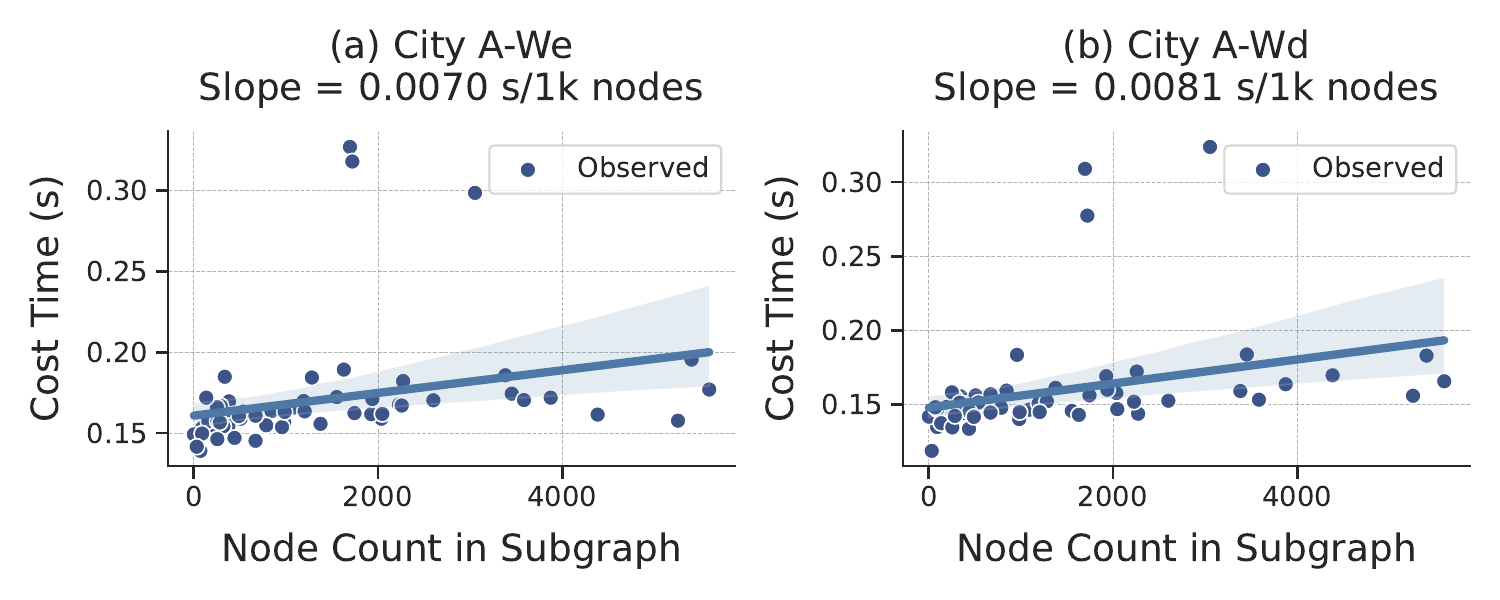}
    \vspace{-10pt}
    \caption{Scalability and inference efficiency of GLIDE.}
    \vspace{-10pt}
    \label{fig:efficiency}
\end{figure}

For large-scale industrial deployment, computational efficiency is as critical as prediction accuracy. We evaluate the training and inference costs of GLIDE against baselines on the City A dataset (fixed scale, no augmentation) using a single NVIDIA Tesla V100 GPU. Standard Transformers (e.g., Informer, Autoformer) exhibit prohibitive training costs due to the quadratic complexity $O(L^2)$ of the self-attention mechanism; Informer requires over 4.5 hours, whereas GLIDE completes in roughly 18 minutes, achieving a {15$\times$ speedup}. Detailed training cost comparison is provided in Appendix~\ref{sec:train_cost}. To verify the scalability of GLIDE for city-wide expansion, we profiled the inference latency with varying input graph sizes ($N$). Figure~\ref{fig:efficiency} reveals a strictly linear scaling law with an extremely low marginal cost: $\text{Slope} \approx 0.007 \text{ seconds} / 1{,}000 \text{ nodes}.$

\subsection{Event-Aware Modulation} 
\label{sec:event-setup-detail}
\subsubsection{Detailed Implementation}

The event knowledge bank is constructed offline via LLM distillation (GPT-4o-mini), covering 123 events across five categories: 19 public holidays, 58 university calendar events, 13 weather events (rainstorms, sustained heat), 30 social events, and 3 platform promotions. Each event is described by its subtype, a five-phase range (buffer\_start / start / middle / end / buffer\_end), and a city-wide or POI-specific spatial scope. For each (subtype, phase, meal-period) tuple, the LLM produces a 16-dimensional impact vector over AOI-type categories. The EventModulator (hidden dim $=128$, 2 layers, dropout $=0.1$) is trained with the backbone frozen using the Adam optimizer ($LR=10^{-3}$, 50 epochs, early stopping patience $=10$). During inference, the modulator applies multiplicative corrections only on positions where the affected time mask $\mathbf{M}_{i,t}=1$, leaving all other predictions identical to the backbone. For event-scenario evaluation, we restrict metrics to these affected positions to isolate the modulator's contribution from regular-period forecasting.

\subsubsection{Event Knowledge Bank Examples}

To make the event-aware modulation pipeline concrete, we provide two representative event records and their corresponding impact vectors from the knowledge bank.

\noindent{\textit{\textbf{Event Record Examples}}}

\noindent\textbf{Example: Weather Event (Rainstorm)}
\begin{center}
\colorbox{gray!10}{%
\begin{minipage}{0.95\columnwidth}
\small\ttfamily
\{\\
\quad "basic\_info": \{\\
\quad\quad "event\_name": "Rainstorm",\\
\quad\quad "event\_type": "weather",\\
\quad\quad "event\_subtype": "rainstorm",\\
\quad\quad "impact\_scope": "city\_wide"\\
\quad\},\\
\quad "data\_description": \{\\
\quad\quad "objective\_info": "Heavy rain Jul 12, 20mm/5min peak",\\
\quad\quad "subjective\_impact": "Demand surge +20-40\%;\\
\quad\quad\quad\quad\quad\quad residential orders rise significantly"\\
\quad\},\\
\quad "time\_info": \{\\
\quad\quad "phase\_ranges": \{\\
\quad\quad\quad "buffer\_start": \{"begin": "07/11", "desc": "forecast"\},\\
\quad\quad\quad "start":      \{"begin": "07/12", "desc": "onset"\},\\
\quad\quad\quad "middle":     \{"begin": "07/12", "desc": "peak"\},\\
\quad\quad\quad "end":        \{"begin": "07/12", "desc": "weaken"\},\\
\quad\quad\quad "buffer\_end":  \{"begin": "07/13", "desc": "recovery"\}\\
\quad\quad\}\\
\quad\},\\
\quad "space\_info": \{"impact\_scope": "city\_wide", "affected\_aoi\_ids": []\}\\
\}
\end{minipage}}
\end{center}

\noindent{\textit{\textbf{Impact Vector Example}}}

For each (subtype, phase, meal-period) tuple, the LLM produces a scalar impact $e \in [-3, +3]$ conditioned on the wave's 16-dimensional AOI-type distribution. Table~\ref{tab:impact_example} shows two illustrative entries.

\begin{table}[h]
\centering
\caption{Illustrative impact vectors from the knowledge bank. The AOI distribution is a 16-dim normalized vector over AOI types: [Residential, Office, University, K-12, Mall, Hospital, Public, Civic, Park, Hotel, Airport, Station, Other, Industrial, Market, Training].}
\label{tab:impact_example}
\small
\begin{tabular}{p{1.2cm}p{0.8cm}p{1.0cm}p{2.5cm}c}
\toprule
\textbf{Subtype} & \textbf{Phase} & \textbf{Meal} & \textbf{Dominant AOI} & \textbf{Scalar} \\
\midrule
New Year & start & lunch & Residential (75\%) & $-3$ \\
New Year & start & lunch & Office+Industrial (53\%) & $-3$ \\
Rainstorm & start & lunch & Residential (100\%) & $+3$ \\
Rainstorm & middle & dinner & Residential (98\%) & $+3$ \\
\bottomrule
\end{tabular}
\end{table}

\noindent\textbf{Interpretation.} For New Year's \textit{start} phase, the LLM assigns $e=-3$ (strong decrease) to office-heavy areas (workplace closures), while for rainstorm \textit{start}, it assigns $e=+3$ (strong increase) to residential areas (residents order delivery during heavy rain). The modulator applies these scalars via the multiplicative correction $\tilde{\mathbf{Q}} = \alpha \odot \hat{\mathbf{Q}} + \beta$, where $\alpha, \beta$ are learned from the temporal modulator inputs derived from $\{e_{i,t}\}$. Events with city-wide scope affect all waves uniformly; POI-specific events affect only waves whose AOI set intersects the event's affected area.

\subsection{Input Reprogramming}
\label{sec:reprogram}

\begin{table}[t]
\centering
\caption{Input Reprogramming study. (a) Look-back window length $T_h$ trade-off: shorter windows reduce latency but degrade accuracy. (b) Symmetric vs. non-symmetric window under the same $T_h=5$ backbone: the symmetric (reprogrammed) window outperforms the non-symmetric one, confirming that input ordering---not just window length or freshness---affects forecast quality. Both panels use the same offline validation split.}
\label{tab:reprogram}
\small
\resizebox{\columnwidth}{!}{
\begin{tabular}{l|ccc|c}
\toprule
\multicolumn{5}{l}{\textit{(a) Look-back window length ($T_h$, symmetric window, pred\_len=10)}} \\
\textbf{Config} & \textbf{CRPS} & \textbf{MAE} & \textbf{WMAPE} & \textbf{Rel. Latency} \\
\midrule
$T_h=3$  & 0.3783 & 0.5029 & 0.5275 & $0.6\times$ \\
$T_h=5$ (symmetric, default) & 0.3494 & 0.4771 & 0.5004 & $1.0\times$ \\
$T_h=10$ & 0.3271 & 0.4535 & 0.4747 & $2.0\times$ \\
$T_h=20$ & \textbf{0.3197} & \textbf{0.4443} & \textbf{0.4639} & $4.0\times$ \\
\midrule
\multicolumn{5}{l}{\textit{(b) Window ordering at $T_h=5$ (same backbone, same val split)}} \\
\textbf{Config} & \textbf{CRPS} & \textbf{MAE} & \textbf{WMAPE} & \textbf{Input Weekday Pattern} \\
\midrule
Non-symmetric (T+7) & 0.3512 & 0.4783 & 0.5043 & [Thu,Fri,Mon,Tue,Wed] \\
\rowcolor{gray!15} \textbf{Symmetric (T+1, reprogrammed)} & \textbf{0.3478} & \textbf{0.4749} & \textbf{0.5007} & [Mon,Tue,Wed,Thu,Fri] \\
\bottomrule
\bottomrule
\end{tabular}
}
\end{table}

As an extension of the event-aware framework, we study how the \emph{organization} of the input window affects the cost-accuracy trade-off---a question of direct operational relevance, since shorter input windows reduce data-collection latency and compute, yet typically degrade accuracy. We view this through the lens of \emph{input reprogramming}: rather than treating the look-back window as a fixed hyperparameter, we explicitly reorganize it (varying its length and the temporal freshness of its samples) to recover accuracy at lower cost. This reprogramming viewpoint echoes the recent trend of reorganizing frozen representations for downstream time-series tasks~\cite{jin2310time,liu2025semantic}, but operates at the level of raw input windows rather than language-model embeddings. Table~\ref{tab:reprogram} reports two controlled studies on the same offline validation split. \textbf{(a) Window length.} We retrain backbone variants with $T_h \in \{3, 5, 10, 20\}$ (default 5) and observe the expected cost-accuracy curve: shortening the window from 10 to 3 degrades CRPS from 0.327 to 0.378 ($+15.7\%$) while reducing the input collection cost by roughly $2\times$. Extending to $T_h=20$ yields further gains (CRPS 0.320) but with diminishing returns and $4\times$ collection cost, confirming that $T_h=5$ offers a practical balance between accuracy and latency. This quantifies the accuracy sacrificed for lower latency and motivates the need for reprogramming to recover it. \textbf{(b) Window ordering.} Under the same $T_h=5$ backbone, we compare a \emph{symmetric} window (input weekday pattern [Mon,Tue,Wed,Thu,Fri] aligned with the target pattern, gap $=1$ day) against a \emph{non-symmetric} window (input pattern [Thu,Fri,Mon,Tue,Wed] cross-week, gap $=7$ days). The symmetric (reprogrammed) window achieves CRPS 0.3478 vs.\ 0.3512 ($-1.0\%$) and MAE 0.4749 vs.\ 0.4783 ($-0.7\%$), demonstrating that aligning input weekdays with target weekdays---without changing the window length---yields a consistent improvement. Note that the symmetric $T_h=5$ result (0.3478) aligns with the default $T_h=5$ in panel (a) (0.3494), confirming cross-panel consistency. Together, these results frame input reorganization---both length and ordering---as a low-overhead lever for balancing efficiency and effect within the event-aware pipeline.

\subsection{Characteristics of Online Production Data}
\label{sec:online_data}
\begin{figure}
    \centering
    \includegraphics[width=1\linewidth]{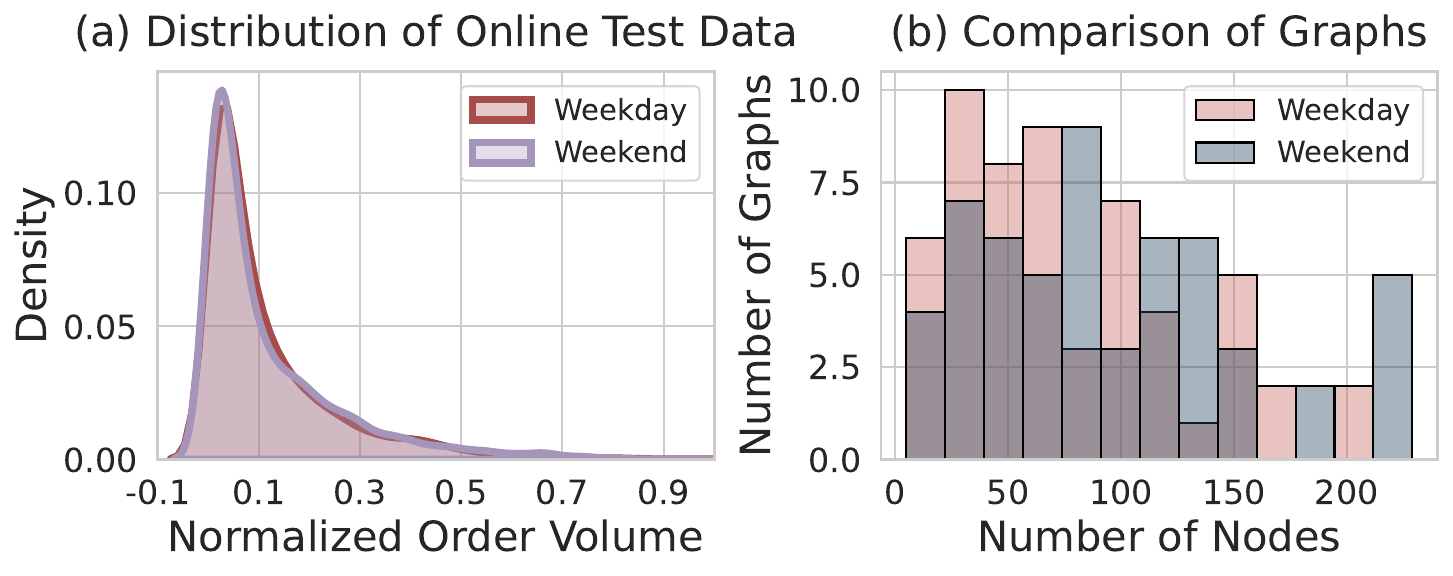}
    \caption{(a) Probability density of normalized order volume for weekdays and weekends. (b) Distribution of subgraph scales (number of nodes) generated during route generation.}
    \label{fig:online_dist}
\end{figure}

Figure~\ref{fig:online_dist} characterizes the online production data in City~A along two axes that directly motivate GLIDE's design. \textbf{(a) Volume distribution.} The normalized order volume follows a heavily right-skewed, zero-inflated distribution: a long tail of low-volume positions coexists with a small set of high-volume bursts, and the weekend distribution is noticeably heavier-tailed than the weekday one, consistent with the bursty, high-entropy lifestyle demand on weekends. This heteroscedasticity motivates both the weekday/weekend split in our benchmarks and the Volume-Adaptive MoE head, which decouples the regression of head and tail regimes rather than collapsing toward the mean. \textbf{(b) Subgraph scale.} The route-planning subgraphs instantiated online are tightly concentrated at small node counts---the vast majority fall well below 300 nodes---with only a thin tail of larger instances. Because these subgraphs are small yet highly variable in size, transductive baselines would require a separately trained model per subgraph; GLIDE's Inductive Graph Quantization instead maps each variable-size subgraph into a fixed set of $K$ latent tokens, keeping the inference cost bounded regardless of $|V_S|$ and enabling real-time deployment on dynamically instantiated inputs.

\subsection{Online Distribution Performance}
\label{sec:online_distribution}
To ensure operational robustness, the system relies on the predicted distribution to assess risk. We evaluate this using Rank-MAE, Prediction Interval Width, Coverage Rate, and Continuous Ranked Probability Score (CRPS). As shown in Table~\ref{tab:online_distribution}, GLIDE demonstrates a decisive advantage in probabilistic modeling: it attains a higher Coverage Rate with a narrower interval Width. For example, in the weekday Critical Interval, GLIDE improves coverage from 76.6\% to 85.3\% while simultaneously reducing the interval width from 9.345 to 7.775. This ``Sharper and Safer'' property implies that GLIDE tightens the uncertainty bounds (lowering wasted reserve capacity) while strictly adhering to safety constraints (avoiding under-supply), directly translating to higher efficiency in the downstream matching system.

\begin{table}[h]
\centering
\caption{Online Distribution Performance. Higher Coverage with lower Width indicates superior uncertainty estimation quality. The ``Critical Interval'' represents the AFUs most sensitive to dispatch optimization.}
\label{tab:online_distribution}
\resizebox{0.48\textwidth}{!}{ 
\begin{tabular}{ll|cccc|cccc}
\toprule
& & \multicolumn{4}{c|}{\textbf{Weekday Scenario}} & \multicolumn{4}{c}{\textbf{Weekend Scenario}} \\
\textbf{Range} & \textbf{Model} & \textbf{R-MAE} & \textbf{Width} $\downarrow$ & \textbf{Cov.} $\uparrow$ & \textbf{CRPS} & \textbf{R-MAE} & \textbf{Width} $\downarrow$ & \textbf{Cov.} $\uparrow$ & \textbf{CRPS} \\ 
\midrule
\multirow{3}{*}{\textbf{$[0, \infty)$}} & Baseline & 2.145 & 6.001 & 81.7\% & 0.386 & 2.302 & 6.320 & 76.2\% & 0.408 \\
& \textbf{GLIDE} & \textbf{1.758} & \textbf{5.014} & \textbf{89.5\%} & \textbf{0.350} & \textbf{2.150} & \textbf{5.875} & \textbf{89.1\%} & \textbf{0.384} \\ 
& \textit{Improv. (\%)} & \textit{+18.04\%} & \textit{+16.45\%} & \textit{+9.55\%} & \textit{+9.33\%} & \textit{+6.60\%} & \textit{+7.04\%} & \textit{+16.93\%} & \textit{+5.88\%} \\
\midrule
\multirow{3}{*}{\textbf{\makecell{Critical\\ Interval}}} & Baseline & 3.556 & 9.345 & 76.6\% & 0.348 & 3.794 & 10.16 & 71.4\% & 0.376 \\
& \textbf{GLIDE} & \textbf{2.918} & \textbf{7.775} & \textbf{85.3\%} & \textbf{0.314} & \textbf{3.503} & \textbf{8.765} & \textbf{86.0\%} & \textbf{0.350} \\ 
& \textit{Improv. (\%)} & \textit{+17.94\%} & \textit{+16.80\%} & \textit{+11.36\%} & \textit{+9.77\%} & \textit{+7.67\%} & \textit{+13.73\%} & \textit{+20.45\%} & \textit{+6.91\%} \\
\bottomrule
\end{tabular}
}
\end{table}

\subsection{Additional Zero-Shot Transfer Results}
\label{sec:zero_shot_extra}
Table~\ref{tab:zero_shot_panel_b} and Table~\ref{tab:zero_shot_panel_c} present the remaining zero-shot transfer pairs that supplement Panel~A in the main paper (Table~\ref{tab:zero_shot}), where GLIDE is trained on City~A and transferred to Cities~B and~C. Here we additionally adopt City~B (medium-scale, 34k nodes) and City~C (small-scale, 19k nodes) as source domains, each transferred to the other two unseen cities without fine-tuning. Compared with Panel~A (Source City~A), transferring from a smaller source domain enlarges the gap to the supervised upper bound, confirming the importance of source-domain diversity for robust zero-shot generalization.

\begin{table}[t!]
\centering
\caption{Panel B: Zero-Shot Transfer from City~B (Medium-scale, 34k nodes). ``GLIDE (Supervised)'' represents the upper-bound performance where the model is trained on the target city itself.}
\label{tab:zero_shot_panel_b}
\resizebox{1.0\columnwidth}{!}{
\begin{tabular}{l|cccc|cccc}
\toprule
\multicolumn{9}{c}{\textbf{Panel B: Source Domain $\rightarrow$ City B (Medium-scale, 34k nodes)}} \\
\midrule
\multirow{2}{*}{\textbf{Model}} & \multicolumn{4}{c|}{\textbf{Target: City A (Large)}} & \multicolumn{4}{c}{\textbf{Target: City C (Small)}} \\
\cmidrule(lr){2-5} \cmidrule(lr){6-9}
 & \multicolumn{2}{c}{Weekday} & \multicolumn{2}{c|}{Weekend} & \multicolumn{2}{c}{Weekday} & \multicolumn{2}{c}{Weekend} \\
 & MAE & RMSE & MAE & RMSE & MAE & RMSE & MAE & RMSE \\
\midrule
HA & 1.335 & 1.895 & 1.455 & 2.091 & 1.186 & 1.689 & 1.182 & 1.649 \\
UniFlow (Zero-shot) & 1.345 & 1.893 & 1.336 & 2.052 & 1.147 & 1.691 & 1.122 & 1.647 \\
\rowcolor{gray!15} \textbf{GLIDE (Zero-shot)} & \textbf{1.147} & \textbf{1.714} & \textbf{1.291} & \textbf{1.947} & \textbf{0.989} & \textbf{1.531} & \textbf{1.006} & \textbf{1.547} \\
\midrule
\textit{Ref: GLIDE (Supervised)} & \textit{1.131} & \textit{1.691} & \textit{1.284} & \textit{1.939} & \textit{0.973} & \textit{1.504} & \textit{0.995} & \textit{1.506} \\
\bottomrule
\end{tabular}
}
\end{table}

\begin{table}[t!]
\centering
\caption{Panel C: Zero-Shot Transfer from City C (Small-scale, 19k nodes).}
\label{tab:zero_shot_panel_c}
\resizebox{1.\columnwidth}{!}{
\begin{tabular}{l|cccc|cccc}
\toprule
\multirow{2}{*}{\textbf{Model}} & \multicolumn{4}{c|}{\textbf{Target: City A (Large)}} & \multicolumn{4}{c}{\textbf{Target: City B (Medium)}} \\
\cmidrule(lr){2-5} \cmidrule(lr){6-9}
 & \multicolumn{2}{c}{Weekday} & \multicolumn{2}{c|}{Weekend} & \multicolumn{2}{c}{Weekday} & \multicolumn{2}{c}{Weekend} \\
 & MAE & RMSE & MAE & RMSE & MAE & RMSE & MAE & RMSE \\
\midrule
HA & 1.335 & 1.895 & 1.455 & 2.091 & 1.246 & 1.781 & 1.293 & 1.892 \\
UniFlow (Zero-shot) & 1.331 & 1.994 & 1.369 & 2.044 & 1.242 & 1.767 & 1.296 & 2.001 \\
\rowcolor{gray!15} \textbf{GLIDE (Zero-shot)} & \textbf{1.166} & \textbf{1.755} & \textbf{1.293} & \textbf{1.955} & \textbf{1.093} & \textbf{1.690} & \textbf{1.140} & \textbf{1.836} \\
\midrule
\textit{Ref: GLIDE (Supervised)} & \textit{1.131} & \textit{1.691} & \textit{1.284} & \textit{1.939} & \textit{1.055} & \textit{1.624} & \textit{1.132} & \textit{1.819} \\
\bottomrule
\end{tabular}
}
\end{table}

\subsection{Cross-City Event Transfer}
\label{sec:cross_city}
To test whether the event modulator generalizes beyond the training city, we transfer the City~A-trained backbone and modulator \emph{zero-shot} to City~B and City~C---two unseen cities with their own locally-collected event libraries. As shown in Table~\ref{tab:event_crosscity}, GLIDE+Event still improves over the GLIDE backbone on CRPS across nearly all event types in both cities. The gains are particularly strong for public holidays (City~B Labor Day $+10.1\%$, Qingming $+7.0\%$; City~C Labor Day $+6.2\%$, Qingming $+5.6\%$), and remain positive for weather events ($+1.0\%$, $+1.1\%$) and platform promotions ($+0.7\%$ to $+1.4\%$). The Qixi anomaly persists ($-0.4\%$, $-0.3\%$), confirming it is a knowledge-bank issue rather than a city-specific one. The positive transfer across two diverse cities confirms that the event-knowledge-bank paradigm---decoupling abstract impact vectors from concrete dates---provides a reusable inductive bias that generalizes across urban logistics networks with different topologies and demand distributions.

\begin{table*}[t!]
\centering
\caption{Cross-city zero-shot transfer to City~B and City~C (unseen during training). The City~A-trained backbone and modulator are applied without retraining. We report affected-sample CRPS and the relative improvement of Phase-2 (GLIDE+Event) over Phase-1 (GLIDE backbone). Positive $\uparrow$ means the event modulator helps.}
\label{tab:event_crosscity}
\resizebox{1.45\columnwidth}{!}{
\begin{tabular}{l|ccc|c|ccc|c}
\toprule
& \multicolumn{4}{c|}{\textbf{City B (34k nodes)}} & \multicolumn{4}{c}{\textbf{City C (19k nodes)}} \\
\textbf{Event} & \textbf{Hist} & \textbf{P1} & \textbf{P2} & \textbf{$\uparrow$} & \textbf{Hist} & \textbf{P1} & \textbf{P2} & \textbf{$\uparrow$} \\
\midrule
\multicolumn{9}{l}{\textit{Weather}} \\
Rainstorm   & 1.4285 & 1.3812 & \textbf{1.3670} & +1.0\% & 1.5849 & 1.5349 & \textbf{1.5184} & +1.1\% \\
Sust. Heat  & --- & --- & --- & --- & 1.6048 & 1.5555 & \textbf{1.5432} & +0.8\% \\
\midrule
\multicolumn{9}{l}{\textit{Platform Promotions}} \\
Double 11   & 1.4222 & 1.3785 & \textbf{1.3683} & +0.7\% & 1.3895 & 1.3513 & \textbf{1.3388} & +0.9\% \\
Double 12   & 1.3752 & 1.3311 & \textbf{1.3118} & +1.4\% & 1.4344 & 1.3841 & \textbf{1.3667} & +1.3\% \\
\midrule
\multicolumn{9}{l}{\textit{Social Events}} \\
Christmas   & 1.3978 & 1.3500 & \textbf{1.3313} & +1.4\% & 1.3948 & 1.3442 & \textbf{1.3266} & +1.3\% \\
Qixi        & 1.4922 & 1.4276 & 1.4339 & -0.4\% & 1.5831 & 1.5215 & 1.5260 & -0.3\% \\
\midrule
\multicolumn{9}{l}{\textit{Public Holidays}} \\
New Year    & 1.7788 & 1.6741 & \textbf{1.5976} & +4.6\% & 1.6567 & 1.5752 & \textbf{1.5721} & +0.2\% \\
Qingming    & 1.6784 & 1.5788 & \textbf{1.4688} & +7.0\% & 1.4882 & 1.4061 & \textbf{1.3273} & +5.6\% \\
Labor Day   & 1.7615 & 1.5960 & \textbf{1.4355} & +10.1\% & 1.5165 & 1.4400 & \textbf{1.3504} & +6.2\% \\
Dragon Boat & 1.4152 & 1.3296 & \textbf{1.3141} & +1.2\% & 1.4303 & 1.3468 & \textbf{1.3208} & +1.9\% \\
\bottomrule
\end{tabular}
}
\end{table*}

\subsection{Ablation on Event Library Coverage}

\label{sec:event_ablation}
A central claim of our event design is that \emph{enriching event types} (beyond legacy public holidays) translates into measurable gains, rather than merely inflating training-time supervision. To verify this, we replace the current 123-event library (covering holidays, university calendars, weather, social, and platform promotions) with a legacy 27-event library (holidays and partial university events only) and re-run inference with the \emph{same} modulator weights. As shown in Table~\ref{tab:event_ablation}, the enriched library increases the affected-sample coverage on every holiday, and---critically---on Dragon Boat the CRPS improvement flips from \emph{negative} (legacy library, $-0.4\%$) to \emph{positive} (enriched library, $+1.3\%$). This is driven by newly added weather events (e.g., a rainstorm on June~22) that raise Dragon Boat coverage from 32.8\% to 100\%, directly demonstrating that event-type expansion makes more samples benefit from modulation rather than being a training-only artifact.

\begin{table}[htbp]
\centering
\caption{Ablation on event library coverage. Replacing the 123-event library with a legacy 27-event library reduces affected-sample coverage and degrades Phase-2 gains. On Dragon Boat, the CRPS improvement of Phase-2 over Phase-1 flips from $-0.4\%$ (legacy) to $+1.3\%$ (enriched), driven by newly added weather events that raise coverage from 32.8\% to 100\%.}
\label{tab:event_ablation}
\small
\resizebox{\columnwidth}{!}{
\begin{tabular}{l|cc|cc|cc}
\toprule
\multirow{2}{*}{\textbf{Holiday}} & \multicolumn{2}{c|}{\textbf{New 123-event}} & \multicolumn{2}{c|}{\textbf{Legacy 27-event}} & \multicolumn{2}{c}{\textbf{Coverage}} \\
 & $N_{\text{aff}}$ & CRPS $\uparrow$ & $N_{\text{aff}}$ & CRPS $\uparrow$ & Legacy & New \\
\midrule
New Year      & 640{,}927 & +1.9\% & 640{,}927 & +1.8\% & 55.9\% & 55.9\% \\
Qingming      & 625{,}165 & +5.0\% & 625{,}165 & +5.0\% & 40.4\% & 40.4\% \\
Labor Day     & 1{,}024{,}930 & +7.3\% & 891{,}855 & +7.8\% & 57.5\% & 66.0\% \\
Dragon Boat   & \textbf{647{,}190} & \textbf{+1.3\%} & \textbf{212{,}370} & \textbf{-0.4\%} & 32.8\% & 100.0\% \\
\midrule
\rowcolor{gray!15} \textbf{All} & \textbf{2{,}938{,}212} & \textbf{+4.0\%} & \textbf{2{,}370{,}317} & \textbf{+4.4\%} & --- & --- \\
\bottomrule
\end{tabular}
}
\end{table}

\subsection{Tail-Robustness Analysis}
\label{sec:tail_robustness}

\begin{table}[h]
\centering
\caption{Performance comparison (MAE) across stratified volume buckets. \textbf{Bold} denotes the best performance, and \underline{underlined} denotes the second best.}
\label{tab:volume_robustness}
\resizebox{1.0\columnwidth}{!}{
\begin{tabular}{l|cccccc}
\toprule
\textbf{Volume Interval} & \textbf{Bucket 1} & \textbf{Bucket 2} & \textbf{Bucket 3} & \textbf{Bucket 4} & \textbf{Bucket 5} & \textbf{Bucket 6} \\
\midrule
HA            & 1.142 & \underline{3.585} & \underline{5.051} & \underline{6.076} & \underline{6.859} & 6.867 \\
w/o MoE       & 1.137 & 3.602 & 5.063 & 6.105 & 7.047 & \underline{6.800} \\
ZIP Baseline  & \underline{1.128} & 3.597 & 5.074 & 6.387 & 6.965 & 6.867 \\
\rowcolor[gray]{0.9} 
\textbf{GLIDE (MoE)} & \textbf{1.106} & \textbf{3.556} & \textbf{5.001} & \textbf{5.937} & \textbf{6.459} & \textbf{6.533} \\
\bottomrule
\end{tabular}
}
\end{table}

To evaluate GLIDE's capability in handling the long-tail demand distribution, we stratify the test set based on the ground-truth order volume. We compare GLIDE against the Historical Average (HA), the ablation variant w/o MoE, and a specialized baseline, the Zero-Inflated Projection (ZIP) method, which explicitly models the probability of zero-counts. Table~\ref{tab:volume_robustness} details the MAE performance across different volume buckets. In high-volume intervals (e.g., Bucket 5), we observe a counter-intuitive phenomenon: the \textit{w/o MoE} neural network (MAE 7.047) actually performs \textit{worse} than the simple Historical Average (MAE 6.859). This highlights a fundamental flaw in using a single regression head for heteroscedastic data. The overwhelming number of zero-value samples dominates the gradient descent, forcing the model to be conservative and bias its predictions towards zero. Consequently, when a high-demand burst occurs, the vanilla model fails to scale up its prediction, resulting in severe under-forecasting errors. GLIDE (with MoE) successfully reverses this trend, achieving the lowest error in the $(40, 50]$ bucket (MAE 6.459), an improvement of roughly 8.3\% over the ablation variant. By employing the \textit{Volume-Adaptive Gating} mechanism, GLIDE identifies high-intensity features in the latent space and routes them to experts specialized in large-value regression. Even compared to the specialized \textit{Zero-Inflated Projection} (MAE 6.965), GLIDE's learnable routing is more effective at distinguishing between true silence and high-volume bursts, demonstrating superior robustness across the entire demand buckets.

\subsection{Training Cost Comparison}
\label{sec:train_cost}

\begin{table}[h]
\centering
\caption{Computational costs on the City A dataset.}
\label{tab:efficiency}
\resizebox{0.7\columnwidth}{!}{
\begin{tabular}{l|cc}
\toprule
\textbf{Model} & \textbf{City A-Wd} & \textbf{City A-We} \\
\midrule
Informer & 4h 33m 35s & 5h 14m 09s \\
STGCN & 3h 22m 45s & 2h 53m 55s \\
Autoformer & 1h 21m 27s & 1h 13m 50s \\
DSTSGNN & 49m 32s & 48m 09s  \\
DLinear & 37m 27s & 30m 33s  \\
STID & 21m 17s & 22m 45s \\
\midrule
\rowcolor{gray!15} \textbf{GLIDE (Ours)} & \textbf{18m 05s} & \textbf{16m 02s}  \\
\bottomrule
\end{tabular}
}
\end{table}

Table~\ref{tab:efficiency} summarizes the complete training time of each model on the City~A dataset. Standard Transformer-based models (Informer, Autoformer) require 1--5 hours, while GLIDE completes training in approximately 18 minutes, yielding a 15$\times$ speedup over the slowest baseline.

\subsection{Case Study: Expert Specialization Visualization}
\label{sec:case_study}
\begin{figure}[h]
    \centering
    \includegraphics[width=1\linewidth]{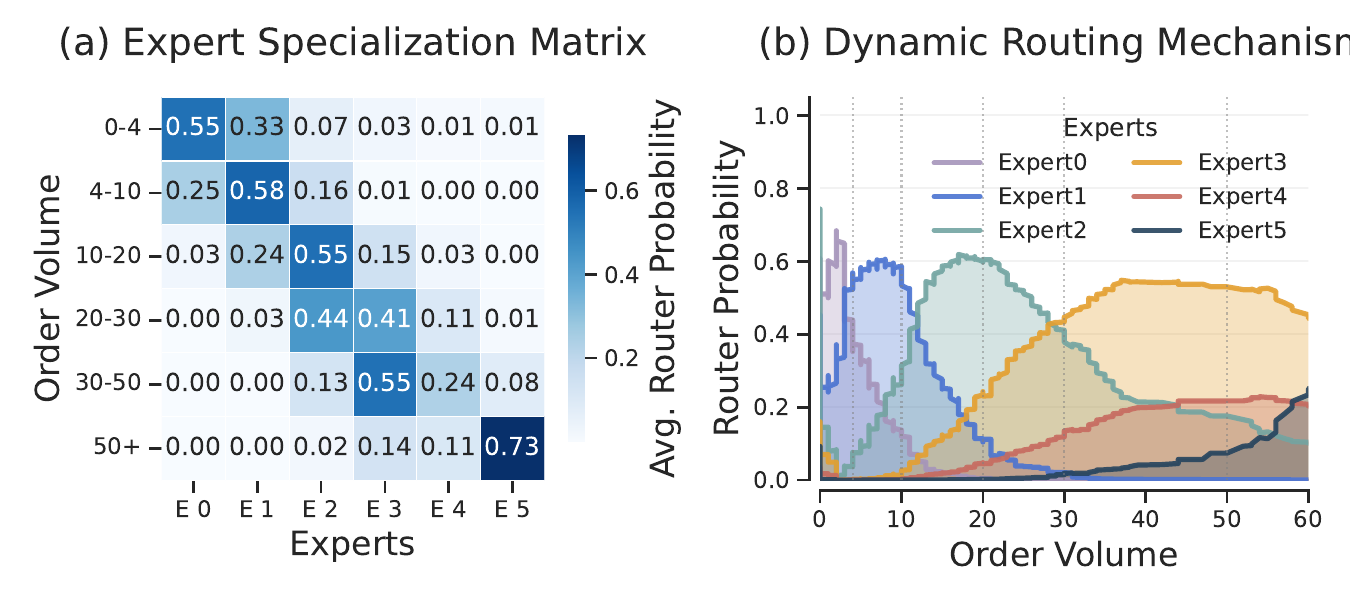}
    \caption{Visualization of the Learned Routing Mechanism on City A (Weekday). (a) Expert Specialization Matrix: The heatmap shows the average activation probability of each expert across discretized volume buckets. (b) Dynamic Routing Mechanism: The curves illustrate the continuous evolution of expert dominance as order volume increases.}
    \label{fig:router_vis}
\end{figure}

To validate that our \textit{Volume-Adaptive MoE} head successfully decouples the learning of different demand regimes, we conducted a visualization case study on the City A-Wd dataset. We extracted the gating probabilities $\pi(x)$ generated by the router network for all test nodes and paired them with their ground-truth order volumes.

\noindent\textbf{Visualization Methodology.}
To construct Figure~\ref{fig:router_vis}, we processed the raw gating signals in two ways. Firstly, we partitioned the test samples into six varying-width intervals (e.g., $[0, 4], [4, 10], \dots, [50, \infty)$) reflecting real-world logistics tiers. We then computed the average routing probability vector for all samples falling into each bin to form the Expert Specialization Matrix. To visualize the dynamic transitions, we sorted all test samples by their ground-truth volume and applied a sliding window smoothing (window size=50) to the router probabilities.

\noindent\textbf{Analysis of Expert Behavior.} The visualization yields two critical insights into the model's inner workings. As evident in Figure~\ref{fig:router_vis}(a), Although individual expert parameters are not directly constrained, the supervised gating objective encourages volume-aware specialization. \textbf{Expert 0 (E0)} acts as a ``Noise Filter,'' dominantly activating (prob 0.55) in the zero-inflated tail $[0, 4]$. Conversely, \textbf{Expert 5 (E5)} acts as a ``Burst Specialist,'' reserving its capacity (prob 0.73) specifically for extremely high-demand events ($>50$ orders). This proves that the model effectively mitigates the ``regression-to-mean'' bias by routing head and tail patterns to distinct parameters. Figure~\ref{fig:router_vis}(b) reveals that the transitions between experts are smooth rather than abrupt. For instance, as the volume increases from 10 to 30, the dominance shifts gracefully from \textbf{E2} (peaking at $\approx 15$) to \textbf{E3} (peaking at $\approx 35$). Notably, in the transition zone of $[20, 30]$, the router assigns comparable weights to both E2 (0.44) and E3 (0.41). This ``Soft Handovers'' mechanism allows the model to blend representations from adjacent experts, maintaining robustness for boundary cases that sit between defined regimes.

\end{document}